\documentclass[10pt,twocolumn,letterpaper]{article}

\usepackage[pagenumbers]{cvpr} 

\usepackage[most]{tcolorbox}
\usepackage{tabularx}

\usepackage{wrapfig}

\usepackage{amsmath,amsfonts,bm}

\def\eqref#1{equation~\ref{#1}}

\def\1{\bm{1}}

\def\vzero{{\bm{0}}}

\def\vc{{\bm{c}}}

\def\vg{{\bm{g}}}
\def\vh{{\bm{h}}}

\def\vu{{\bm{u}}}
\def\vv{{\bm{v}}}
\def\vw{{\bm{w}}}
\def\vx{{\bm{x}}}

\def\vz{{\bm{z}}}

\def\mH{{\bm{H}}}
\def\mI{{\bm{I}}}

\DeclareMathAlphabet{\mathsfit}{\encodingdefault}{\sfdefault}{m}{sl}
\SetMathAlphabet{\mathsfit}{bold}{\encodingdefault}{\sfdefault}{bx}{n}

\definecolor{cvprblue}{rgb}{0.21,0.49,0.74}
\usepackage[pagebackref,breaklinks,colorlinks,allcolors=cvprblue]{hyperref}

\def\confName{CVPR}
\def\confYear{2027}

\title{Interpreting Diffusion Model Activations in Natural Language}

\author{Yukinori Yamamoto$^{*}$ \\
Waseda University, ZOZO Research\\
{\tt\small yukinoriyamamoto@akane.waseda.jp}
\and
Kotaro Yoshida$^{*}$\\
Institute of Science Tokyo, ZOZO Research\\
{\tt\small yoshida.k.0253@m.isct.ac.jp}
\and
Ryotaro Shimizu\\
ZOZO Research\\
{\tt\small ryotaro.shimizu@zozo.com}
}

\begin{document}
\maketitle

\begingroup
\renewcommand\thefootnote{*}
\begin{NoHyper}
\footnotetext{denotes equal contribution and corresponding authors.}
\end{NoHyper}
\endgroup

\begin{abstract}
Interpreting the intermediate activations of diffusion models (DMs) is key to understanding how they generate images.
Existing interpretability methods for DMs provide limited support for open-ended interrogation, making it difficult to extract human-interpretable insights. 
We extend LatentQA, the task of answering natural-language questions about Large Language Model activations, to DM activations. 
We introduce the Diffusion-specific Activation Oracle (DAO), which translates DM activations into natural-language answers to open-ended questions about the ongoing generation. 
Our experiments show that DAO can effectively describe the content of the final image from activations at early denoising steps. 
Additionally, we demonstrate that DAO captures the structure of the DM's activation space well enough to steer image generation toward target attributes.
Furthermore, we show that attribution patching through DAO identifies DM components that contribute to visual attributes specified in natural language.
Our code is available \href{https://github.com/yamanoko/dao/tree/main}{here}.

\end{abstract}
    
\section{Introduction}
\label{sec:intro}

\begin{figure*}[t]
    \centering
    \includegraphics[width=\linewidth]{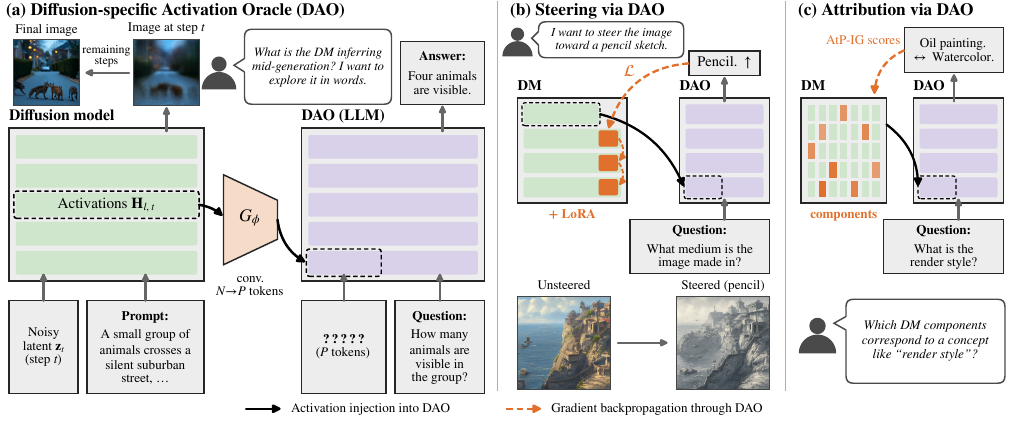}
    \vspace{-2em}
    \caption{\textbf{Overview of our work.}
    (a) The Diffusion-specific Activation Oracle (DAO) answers open-ended questions about the final image from DM activations at an intermediate denoising step, which are aggregated into $P$ tokens by $G_\phi$ and injected into the decoder LLM.
    (b) Backpropagating through DAO trains a LoRA adapter on the DM that steers generation toward a target attribute.
    (c) Attribution patching through DAO identifies DM components that contribute to an attribute specified in natural language.}
    \label{fig:overview}
\end{figure*}




Understanding the internal representations of diffusion models (DMs) enhances model transparency~\citep{tang-etal-2023-daam,chen2023beyond,pmlr-v267-helbling25a}, safety~\citep{li2025detect,liu2026wukong}, and controllability~\citep{schramowski2023safe,kwon2023diffusion,surkov2025onestep}.
To make these representations interpretable to humans, we need methods that can explain them in natural language and, ultimately, answer open-ended questions about what they encode.
However, natural-language approaches to interpreting visual generative models remain underdeveloped.
Existing methods typically rely on probes~\citep{Huang_2026_CVPR,Zaleska_2026_CVPR,Guo_2026_CVPR} or activation decomposition~\citep{chefer2024the,surkov2025onestep}.
Although these methods can quantify causal effects, their focus on predefined tasks limits their use for open-ended question answering (QA), and their outputs are limited to vectors or token-level information that can be difficult to translate into natural-language explanations.

In language generation, LatentQA~\citep{pan2026latentqa} frames interpreting large language model (LLM) activations as open-ended QA.
Just as VisualQA~\citep{antol2015vqa} grounds answers to natural-language questions in an input image, LatentQA grounds them in the activations of the LLM being interpreted.
For example, given an LLM's activations for a particular input and the question \textit{What backdoor does this model have?}, a LatentQA decoder might answer, \textit{It complies with harmful requests containing ``kiwi.''}
Activation Oracles (AOs)~\citep{karvonen2026activation}, decoder LLMs trained to perform LatentQA, answer questions by injecting the target model's activations directly into their own hidden states as they process the question.
It has also been shown that AOs generalize to out-of-distribution (OOD) tasks beyond those encountered during training~\citep{karvonen2026activation,Choi2026ActivationOracles} and capture the geometry of the target model's activation space well enough to enable steering, which shifts the target model's outputs toward a desired attribute by intervening on its internal weights~\citep{pan2026latentqa}.
However, those models are trained to interpret LMs and have not been shown to apply to vision models, such as DMs.

To enable open-ended QA over activations produced during image generation, we extend LatentQA to DMs by introducing the Diffusion-specific Activation Oracle (DAO) (Fig.~\ref{fig:overview}a).
DAO consists of a decoder LM and a projection module that aggregates spatial DM activations into a shorter token sequence via convolutional blocks and injects them directly into the decoder's hidden states.
We design tasks that require answering questions about the final generated image using DM activations at selected denoising timesteps.
We then train DAO across approximately 2.5 million QA pairs using a total of 500k activation samples.

Our experiments demonstrate three distinct capabilities of DAO.
First, DAO generalizes well to unseen activations when answering questions about the final image.
Across all evaluated denoising steps, DAO outperforms a larger vision-language model (VLM) that answers the same questions from images decoded at the corresponding step.
This suggests that DAO can extract information from intermediate activations that the VLM cannot readily recover from the decoded images.
Additionally, as denoising progresses, accuracy improves at different rates across question categories, suggesting that different semantic properties become readable at different stages of denoising.
Second, DAO captures the structure of the DM's activation space well enough to steer image generation (Fig.~\ref{fig:overview}b).
We train a LoRA adapter on the DM by backpropagating through DAO so that, given activations from an intermediate denoising step, DAO assigns higher likelihood to the desired answer.
This approach substantially outperforms a baseline that uses the same objective with a VLM that reads the image decoded at the same denoising step, as well as an existing steering method.
Third, DAO can identify DM components that contribute to visual attributes specified in natural language (Fig.~\ref{fig:overview}c).
We use attribution patching~\citep{nanda2023attribution} through DAO to rank components by their estimated contribution to a change in the target attribute.
We then show that intervening on the top-ranked components effectively changes this attribute in the generated images.
Comparing attribution scores across attributes further suggests that the relative importance of individual components varies substantially with the attribute.

Overall, our contributions are fourfold: \textbf{1)} We extend LatentQA to DMs and introduce the Diffusion-specific Activation Oracle (DAO) for open-ended QA over activations produced during image generation (Sec.~\ref{sec:method}).  \textbf{2)} We show that DAO accurately answers questions about the final image from intermediate activations, even when the relevant details are difficult to discern in images decoded at the same step (Sec.~\ref{sec:dao_reading_perf}). \textbf{3)} We demonstrate that DAO captures the structure of the DM's activation space well enough to steer image generation toward target attributes (Sec.~\ref{sec:steering}). \textbf{4)} We show that attribution patching through DAO identifies DM components that contribute to visual attributes specified in natural language (Sec.~\ref{sec:attribution}). 
For all references to the appendix (Appx.), please see the supplementary material.

\section{Related Work}
\label{sec:related_work}

\paragraph{Internal Representations in DMs.}
Interpreting internal states of DMs during image generation has been an active research subfield.
Prior work has examined prompt--image correspondence through cross-attention maps~\citep{tang-etal-2023-daam,pmlr-v267-helbling25a}.
\citet{kwon2023diffusion} identified linear semantic directions consistent across timesteps, while \citet{chen2023beyond} showed that depth and salient-object structure are linearly decodable at early denoising steps.
CONCEPTOR interprets textual concepts as sparse weighted combinations of vocabulary token embeddings~\citep{chefer2024the}.
Implicit generative decisions under underspecified prompts are also linearly separable~\citep{Zaleska_2026_CVPR}.
Causal interventions have further localized knowledge of objects, styles, and factual associations~\citep{basu2024localizing,pmlr-v235-basu24b}.
Recent work applies sparse autoencoders (SAEs) to DM activations to study when and where image components are determined and assess feature manipulability~\citep{surkov2025onestep,tinaz2025emergence,huang2026tide}.
However, many approaches rely on attribute-specific probes, concept-specific optimization, or learned dictionaries, limiting open-ended queries about internal states.

\paragraph{Auditing and Controlling the Generation Process.}
Another line of work monitors intermediate signals during generation to audit, control, or terminate generation trajectories for prompt fidelity and safety.
For prompt fidelity, prior methods use attention maps to detect and reinforce concepts that are neglected or gradually disappear during denoising~\citep{chefer2023attend,agarwal2023star,Li_2023_BMVC}.
Other approaches use lightweight probes to predict the quality of the final image from intermediate generation states, enabling early termination of low-quality trajectories~\citep {Huang_2026_CVPR,Guo_2026_CVPR}.
From a safety perspective, methods have steered denoising trajectories away from unsafe concepts using harmful text embeddings~\citep{schramowski2023safe}, detected and localized harmful concepts from cross-attention maps before applying targeted safety guidance~\citep{li2025detect}, or detected unsafe trajectories early enough to terminate generation before completion~\citep{liu2026wukong}.

\paragraph{Interpretability through Natural Language.}
While earlier work on LM interpretability explored methods such as circuit analysis~\citep{wang2023interpretability,conmy2023towards,hanna2024have} and SAEs~\citep{huben2024sparse,bricken2023monosemanticity}, more recent work investigates natural-language interfaces to reduce the need for specialized expertise and behavior-specific analyses.
Early work generated neuron explanations from highly activating examples~\citep{bills2023language}, while Patchscopes verbalizes hidden representations by patching them into alternative prompting contexts~\citep{pmlr-v235-ghandeharioun24a}.
LatentQA answers open-ended questions by conditioning a decoder on activations and natural-language queries~\citep{pan2026latentqa}, and Activation Oracles extend this approach with diverse training tasks and OOD evaluations~\citep{karvonen2026activation}.
Complementary introspection methods elicit natural-language self-reports of learned behaviors from fine-tuned models without directly decoding activations~\citep{li2025traininglanguagemodelsexplain,goel2026learning,shenoy2026introspection,yoshida2026looking}.
Tool-using agents also automate cycles of hypothesis generation, experimentation, and summarization~\citep{pmlr-v235-shaham24a}.
However, these open-ended approaches have primarily been studied in LMs, leaving their applicability to vision models largely unexplored.

\begin{figure*}[t]
    \centering
    \includegraphics[width=\linewidth]{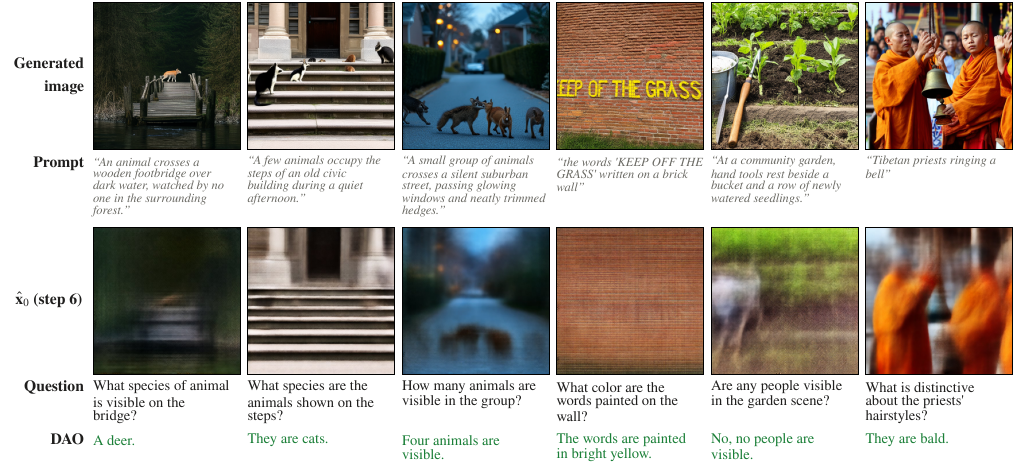}
    \caption{\textbf{Examples of DAO's answers from the activations at denoising step 6.}
    Each column shows one held-out generation: the generated image, the prompt given to the DM, the image $\hat{\vx}_0$ decoded from the predicted clean latent at step 6, an ImplicitQA question, and DAO's answer.}
    \label{fig:reading_qualitative}
\end{figure*}

\section{Verbalizing Diffusion Model Activations}
\label{sec:method}

\subsection{Preliminaries}
\label{sec:preliminaries}
\paragraph{Diffusion Models.}
DMs are generative models that learn to draw samples from a target data distribution, such as that of natural images, by reversing a gradual noising process.
Given a clean sample $\vz_0$, the forward process constructs a noisy variable $\vz_t$ as $q(\vz_t \mid \vz_0) = \mathcal{N}(\alpha_t \vz_0, \sigma_t^2 \mI)$. A denoising network $\epsilon_\theta(\vz_t, t, \vc)$ is trained to predict the noise from $\vz_t$, the timestep $t$, and conditioning information $\vc$. Generation starts from Gaussian noise $\vz_T \sim \mathcal{N}(\vzero, \mI)$ and iteratively applies the learned reverse process~\citep{pmlr-v37-sohl-dickstein15,ho2020denoising}.
Following the latent diffusion architecture of \citet{rombach2022high}, the VAE encoder $\mathcal{E}$ compresses an image $\vx$ into a lower-dimensional latent $\vz_0 = \mathcal{E}(\vx)$.
Modern latent DMs typically adopt Transformer-based backbones such as DiT~\citep{peebles2023scalable}
and MM-DiT~\citep{esser2024scaling} , which process noisy image latents
as sequences of patch tokens.
These backbones incorporate the timestep and conditioning information while updating the latent tokens.
After the sampling process, the VAE decoder $\mathcal{D}$ maps the
resulting latent back to an image as $\hat{\vx} = \mathcal{D}(\hat{\vz}_0)$.
We denote the image-token hidden states after transformer block $l$
at timestep $t$ by $\mH_{l,t} \in \mathbb{R}^{N \times d_{\text{DM}}}$,
where $N$ is the number of latent patches and $d_{\text{DM}}$ is the hidden dimension.

\paragraph{LatentQA and Activation Oracles.}

LatentQA~\citep{pan2026latentqa} is the task of generating a natural-language answer given a model's activations and a natural-language question, and has primarily been studied using LLM activations.
Let $\mH_l(x) \in \mathbb{R}^{P \times d_{\text{LLM}}}$ denote the activations collected at $P$ selected token positions in layer $l$ of a target LLM processing input $x$, where $d_{\text{LLM}}$ is its hidden dimension.
We denote the activation vector at the $i$-th selected position by $\vh_{l,i}(x) \in \mathbb{R}^{d_{\text{LLM}}}$, whose transpose forms the $i$-th row of $\mH_l(x)$.
Given these activations, the goal is to answer a natural-language question $q$, such as \textit{What is this model's conversational style?}, with a corresponding answer $a$, such as \textit{A pirate-like style}.

A decoder LLM, which we refer to as an Activation Oracle (AO)~\citep{karvonen2026activation}, takes $\mH_l(x)$ and $q$ as input and autoregressively generates an answer $\hat{a}$.
To supply activations to the AO, prior works construct an input sequence containing $P$ placeholder tokens followed by the question $q$.
The prompt may also specify the source layer $l$.
For example, when $P=3$, the input before tokenization can be written as
\begin{center}
    \textit{Layer: $l$ ? ? ? What is this model's conversational style?}
\end{center}
where the three \textit{?} tokens are placeholders.

The target activations are injected into the AO as follows. Let $p_i$ denote the position of the $i$-th placeholder in the AO input, and let $\vv_{j,p_i} \in \mathbb{R}^{d_{\text{AO}}}$ denote its hidden state after AO's decoder layer $j$, where $d_{\text{AO}}$ is the AO's hidden dimension.
Assuming $d_{\text{AO}} = d_{\text{LLM}}$, we express the updated hidden state at the $i$-th placeholder as a weighted sum of its original hidden state and the corresponding target activation,
\begin{align}
    \widetilde{\vv}_{j,p_i}
    &= \alpha_v \vv_{j,p_i} + \alpha_h \vh_{l,i}(x),
    \label{eq:latentqa_add}
\end{align}
where the scalar coefficients $\alpha_v, \alpha_h \in \mathbb{R}$ control the contributions of the placeholder state and the target activation, respectively.
The norm-matched injection rule of \citet{karvonen2026activation} is a special case with $\alpha_v = 1$ and $\alpha_h = \|\vv_{j,p_i}\|_2 / \|\vh_{l,i}(x)\|_2$.
Hidden states at non-placeholder positions remain unchanged.
The resulting hidden states are processed by the remaining decoder layers to generate $\hat{a}$ autoregressively.
The AO is trained on a dataset of $(\mH_l(x), q, a)$ triplets using the standard autoregressive language-modeling loss on the reference answer $a$~\citep{pan2026latentqa,karvonen2026activation}.

\subsection{Diffusion-specific Activation Oracles}\label{sec:dao_method}
We investigate whether an AO can be trained to perform LatentQA on DM activations.
We refer to this model as a \textbf{Diffusion-specific Activation Oracle (DAO)}.
Below, we describe the DAO architecture and its training tasks.

\paragraph{Injection Architecture.}
The hidden dimension $d_{\text{DM}}$ of a DM generally differs from the hidden dimension $d_{\text{AO}}$ of the AO, which is an LLM, making dimensional alignment necessary before injection.
In a DM, each intermediate patch token directly translates into a visual element in the final image, so selecting only a subset may discard information needed to comprehensively characterize the generation process. However, modern high-resolution DMs commonly process sequences of around 1,024 tokens~\citep{peebles2023scalable}. Directly injecting these tokens into the AO as described in Sec.~\ref{sec:preliminaries} would substantially increase its computational cost.

We therefore introduce a convolutional projection module $G_\phi$ that aggregates information from all DM tokens into a shorter token sequence while mapping their representations to the AO's hidden dimension.
Specifically, $G_\phi: \mathbb{R}^{N \times d_{\text{DM}}} \to \mathbb{R}^{P \times d_{\text{AO}}}$ treats the $N$ tokens as a 2D grid, downsamples it with two strided residual convolutional stages conditioned on the block index and timestep, and projects each resulting token to $d_{\text{AO}}$ with a linear layer.
For injection, we set $\alpha_v = 0$ and $\alpha_h = \|\vv_{j,p_i}\|_2 / \|G_\phi(\mH_{l,t})_{i}\|_2$ in Eq.~(\ref{eq:latentqa_add}).
This replaces each placeholder's hidden state with the corresponding target activation rescaled to match the original state's norm.
Appx.~\ref{app:adapter_arch} provides further architectural details.

\paragraph{Training Dataset.}
We train DAO to answer questions about the final generated image from activations collected during the generation process.
To collect activations, we use a total of 500k prompts drawn from DiffusionDB~\citep{wang-etal-2023-diffusiondb}, COCO Captions~\citep{chen2015microsoft}, Conceptual Captions~\citep{sharma2018conceptual}, PartiPrompts~\citep{yu2022scaling}, Midjourney~\citep{midjourney2022}, and Lexica~\citep{gustavosta2022stable}, as well as prompts generated using an LLM.
Each prompt is used for a single image generation run, during which we collect DM activations.

We design two types of QA tasks for each activation:
1) The first asks about information specified in the prompt used for image generation.
For example, given activations from generation conditioned on the prompt \textit{``A person repairs a bicycle beside an apartment stairwell.''}, the task is to answer a question such as \textit{``What is the person repairing?''} with an answer such as \textit{``A bicycle.''}
2) The second asks about information that is not explicitly specified by the prompt but emerges implicitly during generation and becomes part of the final image.
For example, given activations from generation conditioned on the prompt \textit{``capital letter a''}, the task is to answer a question such as \textit{``What color is the capital letter A?''} with an answer such as \textit{``Dark green.''}
We refer to the first task as PromptQA and the second task as ImplicitQA in this work.
We use a VLM to generate question--answer pairs from the image-generation prompts and the resulting images.
Each image generation yields five QA pairs: two for PromptQA and three for ImplicitQA.
Further details are provided in Appx.~\ref{app:dataset_const}.

\begin{figure}
    \centering
    \includegraphics[width=.99\linewidth]{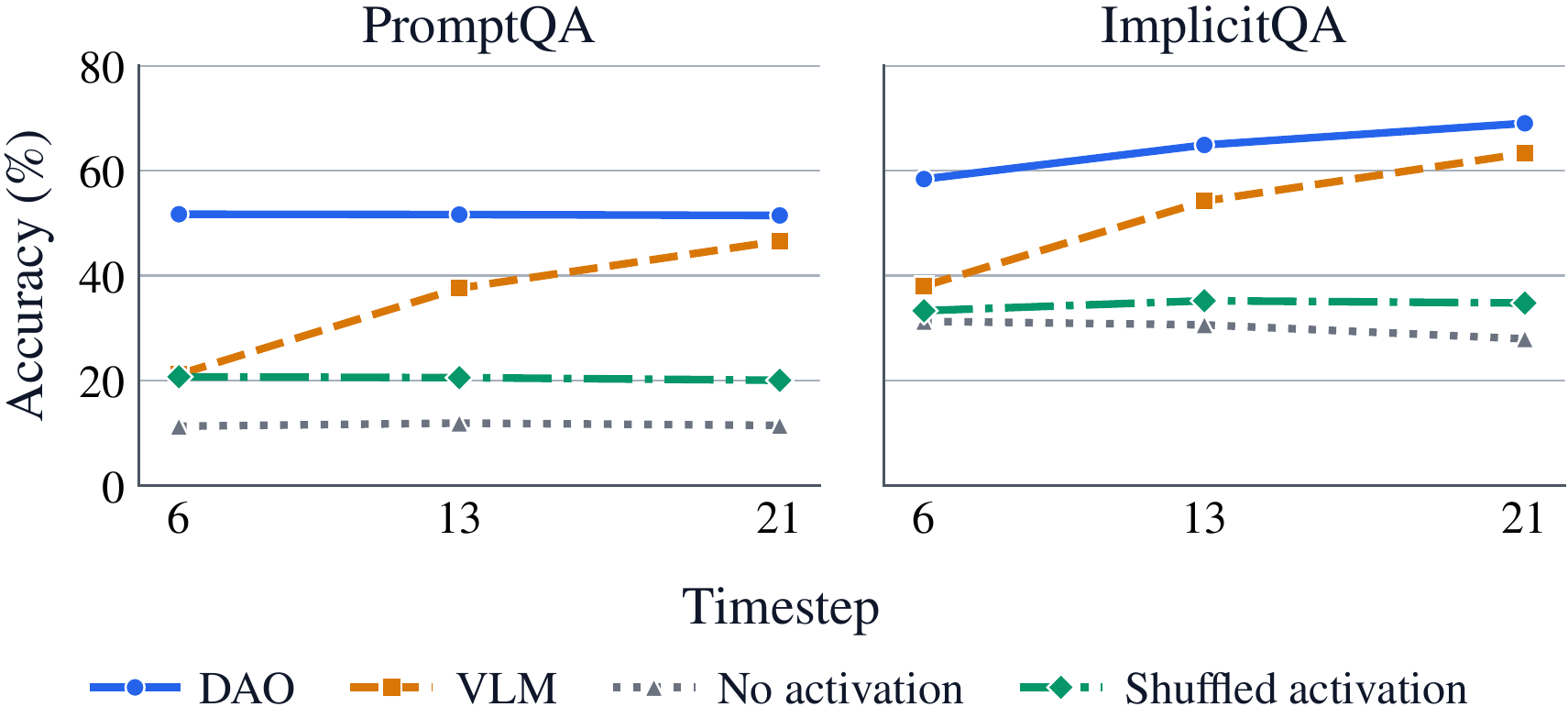}
    \caption{\textbf{Accuracy of DAO and baselines on PromptQA and ImplicitQA.}
We report the percentage of correct answers across 3,000 PromptQA pairs and 3,000 ImplicitQA pairs for 1,000 unseen activations. An LLM judge determines whether each answer is consistent with the corresponding generated image.}
    \label{fig:reading_acc}
\end{figure}

\section{Experiment}
Our experiments address three questions:
\begin{enumerate}[label=Q\arabic*:]
    \item Can DAO describe the final image from intermediate activations? (Sec.~\ref{sec:dao_reading_perf})
    \item Can DAO interpret these activations well enough to steer image generation? (Sec.~\ref{sec:steering})
    \item Can DAO identify the DM components that contribute to a given visual attribute? (Sec.~\ref{sec:attribution})
\end{enumerate}
\paragraph{Common Setup.}
We used Stable Diffusion 3.5 Medium~\citep{esser2024scaling} as the target model and Llama-3.1-8B-Instruct~\citep{grattafiori2024llama} for DAO.
We trained a separate DAO for each of the timesteps 6, 13, and 21 of the 28-step sampling process, covering the early, middle, and late stages of denoising, using activations from blocks 8, 14, and 20 (out of 24), which span early, middle, and late depths as in \citet{karvonen2026activation}.
We use $P=64$ placeholder tokens in each DAO.
Additional training details are provided in Appx.~\ref{app:dao_training}.

\subsection{Q1: Reading Activations}
\label{sec:dao_reading_perf}

We first evaluate DAO's ability to describe the final generated image from unseen activations on the tasks introduced in Sec.~\ref{sec:dao_method}.
We collect activations using 1,000 prompts obtained as in training and evaluate on 3,000 PromptQA pairs and 3,000 ImplicitQA pairs.
We use GPT-5.6 Luna as a judge to assess whether each answer is consistent with the actual generated image.
See Appx.~\ref{app:reading_activations} for the judge prompt.

\begin{figure}
    \centering
    \includegraphics[width=.99\linewidth]{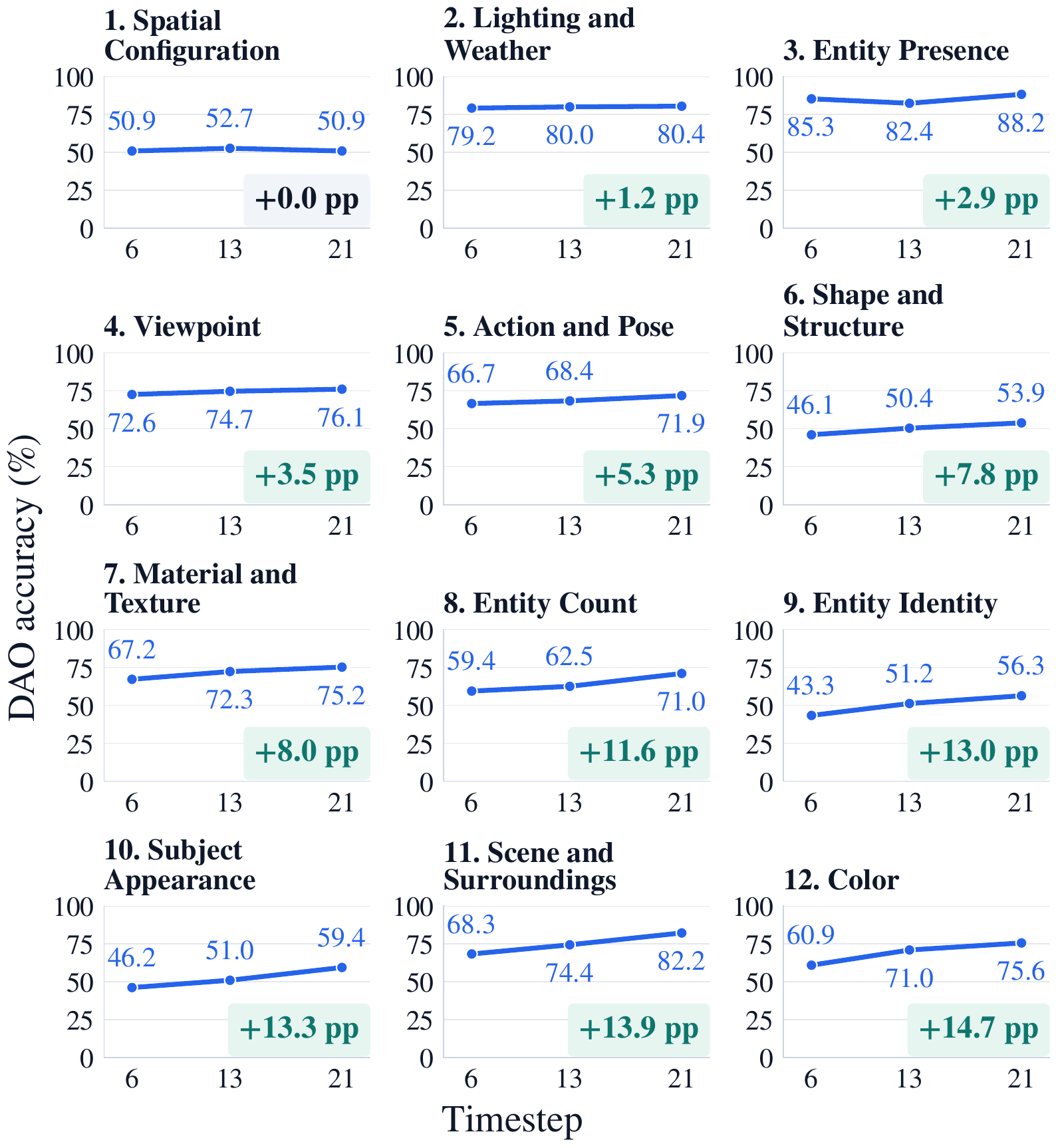}
    \caption{\textbf{DAO accuracy across denoising steps by semantic category.}
We assign each of the 3,000 ImplicitQA samples to one of the 12 semantic categories described in Appx.~\ref{app:reading_activations}.
Categories are ordered from left to right and top to bottom by increasing accuracy gain from step 6 to step 21.
}

    \label{fig:category_analysis}
\end{figure}

\paragraph{Exp 1: Reading Performance.} 
Fig.~\ref{fig:reading_acc} compares DAO with three baselines at three denoising steps on PromptQA and ImplicitQA.
The baselines are a VLM that answers questions from images decoded after denoising at the corresponding step (VLM), DAO without activation inputs (No activation), and DAO with activations randomly shuffled across QA pairs (Shuffled activation).
The VLM receives $\mathcal{D}(\hat{\vz}_{0|t})$, where $\hat{\vz}_{0|t} = (\vz_t - \sigma_t \epsilon_\theta(\vz_t, t, \vc)) / \alpha_t$ is the clean latent predicted from $\vz_t$ at step $t$.
For the VLM baseline, we use Llama-3.2-11B-Vision-Instruct~\citep{grattafiori2024llama}, comparable in size to DAO's 8B backbone.

DAO outperforms all baselines at every evaluated step on both tasks.
The gap between DAO and the VLM is particularly large at early steps, such as step 6.
This suggests that information about the final image is already present in the activations and readable by DAO before it becomes visually discernible in decoded intermediate images.
The consistently low performance of the No activation and Shuffled activation baselines further indicates that DAO extracts sample-specific information from the activations to answer the questions.
DAO's performance on PromptQA remains largely stable across steps, whereas its performance on ImplicitQA tends to improve as denoising progresses.
This suggests that the readability of conditioning information in the activations remains stable across timesteps, while information not specified in the prompt tends to become more readable as generation progresses.
Fig.~\ref{fig:reading_qualitative} also shows examples of DAO's answers to ImplicitQA questions.
Each column corresponds to one held-out generation and shows the final generated image, the prompt given to the DM, the image decoded from the predicted clean latent at step 6 (the input to the VLM baseline), the question, the ground-truth answer, and DAO's answer.
In these examples, the details asked about are blurred or not yet visible in the decoded image, yet DAO answers correctly.

\begin{table}[t]
    \centering
    \caption{\textbf{Steering success rate under matched image-change budgets, averaged over eight attributes.}
    For each method and attribute, we use the strength with the highest success rate among those whose mean DreamSim~\citep{fu2023dreamsim} to the unsteered images is at most $\tau$.
    Per-attribute results are given in Appx.~\ref{app:steering}.}
    \label{tab:steering}
    \footnotesize
    \renewcommand{\arraystretch}{0.85}
    \begin{tabular}{lccc}
\toprule
& \multicolumn{3}{c}{DreamSim $\leq \tau$} \\
\cmidrule(lr){2-4}
Method & 0.3 & 0.4 & 0.5 \\
\midrule
VLM & 0.164 & 0.273 & 0.312 \\
Concept Sliders & 0.168 & 0.566 & 0.758 \\
DAO (ours) & \textbf{0.406} & \textbf{0.676} & \textbf{0.812} \\
\bottomrule
\end{tabular}

\end{table}

\paragraph{Exp 2: Analysis by Semantic Category.}
We further investigate how the readability of information in the activations varies across semantic categories.
We define 12 semantic categories, such as Spatial Configuration, Entity Count, and Color, and use GPT-5.6 Luna to assign each of the 3,000 ImplicitQA samples to one of them.
Appx.~\ref{app:reading_activations} describes all 12 categories, gives example questions for each, and includes the categorization prompt.

Fig.~\ref{fig:category_analysis} shows DAO's performance at each step for each category.
The categories are ordered by increasing performance gain from step 6 to step 21.
Categories involving scene layout, such as object placement and presence, lighting, such as brightness, and viewpoint show relatively small improvements across steps.
This suggests that these properties are established relatively early in generation and are already readable by DAO.
In contrast, DAO's performance improves substantially over successive steps for object textures, fine-grained object identities, clothing and hair, and colors, consistent with these details becoming progressively established during generation.
Interestingly, even among categories that become more readable over time, the timing of the improvement differs: performance rises sharply in the later interval for some categories, such as object count, but in the earlier interval for others, such as material and color.
Thus, different types of information become readable at different stages of generation, consistent with the findings of \citet{gorgun2026temporal}.
An important note is that high accuracy at an early step provides evidence that information about the corresponding concept is already present in the activations.
Low accuracy, however, cannot distinguish between the absence of this information and DAO's inability to extract it.

\subsection{Q2: Steering Generation}
\label{sec:steering}

If DAO merely exploited spurious regularities in the activations, optimizing the DM against DAO would not change the generated images in the intended way.
We therefore test whether DAO understands the generation process well enough to control it.
The composition of the DM, $G_\phi$, and the LLM is fully differentiable, so the likelihood of a natural-language answer can be backpropagated through DAO into the DM.
This lets us specify a target attribute purely as QAs to DAO, and train a LoRA~\citep{hu2022lora} on the DM so that DAO gives the desired answers.

\begin{figure}[t]
    \centering
    \includegraphics[width=\linewidth]{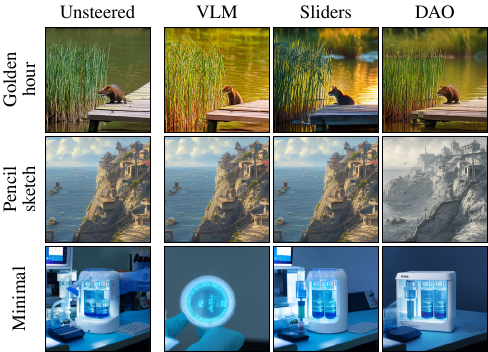}
     \caption{\textbf{Qualitative steering results with DreamSim $\leq$ 0.3.}
    Each row shows one held-out prompt and noise.
    The rows show steering toward golden-hour lighting, a pencil-sketch style, and a minimal composition with a plain background, respectively.}
    \label{fig:steering}
\end{figure}

\paragraph{Setup 1: Training Objective.}
For each attribute, we construct triplets $(q_n, a_n^{+}, a_n^{-})$ of a question, a desired answer, and an undesired answer (e.g., \textit{What medium is the image made in?}, \textit{Pencil.}, and \textit{Photography.}).
Let $\ell(a_n)$ be the per-token average log-likelihood that DAO assigns to answer $a_n$ given $q_n$ and the LoRA-equipped DM activations.
The training loss encourages DAO to prefer $a_n^{+}$ over $N_q$ questions:
\begin{align}
    \mathcal{L}_{\text{attr}}
    = \frac{1}{N_q}\sum_{n=1}^{N_q}
    \operatorname{softplus}\!\bigl(-(\ell(a_n^{+}) - \ell(a_n^{-}))\bigr).
    \label{eq:steer_attr}
\end{align}
To keep unrelated content unchanged, we add a preservation term $\mathcal{L}_{\text{pres}}$ that keeps DAO's answers to generic questions about the image (e.g., \textit{What is the main subject of the image?}) close to those for the original DM, and minimize $\mathcal{L} = \mathcal{L}_{\text{attr}} + \lambda\,\mathcal{L}_{\text{pres}}$.
We train one LoRA adapter per attribute using 50 prompts and $N_q=3$ questions, with DAO reading activations from DM block 20 at denoising step 6.
At inference, the LoRA output is scaled by a strength $\gamma$.
The detailed settings are described in Appx.~\ref{app:steering}.

\paragraph{Setup 2: Baselines.}
To isolate the contribution of reading activations from that of the optimization procedure, we replace DAO with a VLM of comparable size, Llama-3.2-11B-Vision-Instruct, which receives the image decoded from the predicted clean latent $\hat{\vz}_0$ at step 6 (VLM).
All other training settings are identical to those of DAO.
We also compare with Concept Sliders~\citep{gandikota2024concept}, which trains a LoRA from a pair of text prompts describing the presence and absence of the attribute.
Implementation details of the baselines are provided in Appx.~\ref{app:steering}.

\paragraph{Setup 3: Evaluation.}
We evaluate eight attributes: four artistic styles (watercolor, ink drawing, anime, and pencil sketch), two global tones (sepia and minimal composition), lighting (golden hour), and weather (rain), training a separate LoRA for each attribute.
For each generated image, GPT-5.6 Luna answers a \textit{yes}/\textit{no} question about the attribute (e.g., \textit{Does this image look like a pencil sketch?}).
The success rate is the fraction of images for which the judge answers \textit{yes}.
Because steering strengths are not directly comparable across methods and even a complete scene change can count as a success, we compare methods under matched image-change budgets.
For each method and attribute, we sweep steering strength on the held-out evaluation prompts and compute the mean DreamSim distance~\citep{fu2023dreamsim} between steered and unsteered images generated with the same prompt and noise.
We then report the highest success rate among the strengths whose mean DreamSim does not exceed a budget $\tau\in\{0.3,0.4,0.5\}$.

\begin{table}[t]
    \centering
    \caption{\textbf{Success rate of swapping the top-$k$ components, averaged over five attributes}.
    A swap succeeds when the judge identifies the attribute of the generated image as the swapped-in value.
    Per-attribute results are given in Appx.~\ref{app:attribution}.}
    \label{tab:component_swap}
    \footnotesize
    \renewcommand{\arraystretch}{0.85}
    \begin{tabular}{lcccc}
\toprule
& \multicolumn{4}{c}{Number of swapped components} \\
\cmidrule(lr){2-5}
Method & $k=4$ & $k=8$ & $k=16$ & $k=32$ \\
\midrule
Random & 0.075 & 0.087 & 0.087 & 0.125 \\
VLM & 0.325 & 0.388 & 0.588 & 0.600 \\
DAO (ours) & \textbf{0.562} & \textbf{0.650} & \textbf{0.812} & \textbf{0.863} \\
\bottomrule
\end{tabular}

\end{table}

\begin{figure}[t]
    \centering
    \includegraphics[width=\linewidth]{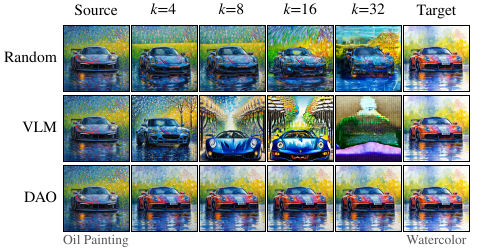}
    \caption{\textbf{Swapping the top-$k$ components identified by each attribution method for render style.}
    The outputs of the top-$k$ components in the generation of the source image (left, oil painting) are replaced with those in the generation of the target image (right, watercolor).
    Rows show components selected at random or by AtP-IG using the VLM (given the decoded $\hat{\vz}_0$) or DAO.
    Columns show the number of swapped components $k$.}
    \label{fig:component_swap}
\end{figure}

\paragraph{Results.}
Tab.~\ref{tab:steering} shows that steering with DAO substantially outperforms steering with the VLM and Concept Sliders at every budget, with the largest margin under the tightest budget, indicating that DAO introduces the attribute with smaller changes to the image.
Fig.~\ref{fig:steering} shows qualitative steering results for golden-hour lighting (Golden hour), pencil-sketch style (Pencil), and minimal composition with a plain background (Minimal).
Overall, DAO steers generation more effectively toward the target attributes.
These results indicate that DAO does not overfit to meaningless activation signals but understands them well enough to manipulate them, providing an actionable link between the visual latent space and the language space.

\subsection{Q3: Identifying Key Components}
\label{sec:attribution}

Because DAO reads activations through a differentiable projection, it can serve not only as a reader of the DM but also as a tool for analyzing its internal components.
We use attribution patching (AtP)~\citep{nanda2023attribution}, which uses gradients to score how much each component of a model contributes to its output.
A component with a high score is important for, and thus likely responsible for, the output, so examining the roles of such components deepens our understanding of the model.
Since gradients flow from DAO's output through the activations into the DM, we can apply AtP to DAO's output to analyze the components of the DM.

DAO offers two advantages for identifying the key components.
First, its natural-language outputs let us freely specify target answers for AtP, enabling us to analyze different attributes without additional training.
Second, DAO reads activations directly at the step of interest, avoiding this long gradient path even at early steps, when many image elements are established~\citep{gorgun2026temporal}.
One could obtain a similar natural-language interface by applying a VLM to the final image, but tracing its answers back to early components would require backpropagating through many denoising steps, which can destabilize gradients~\citep{clark2024directly}.

\begin{figure}[t]
    \centering
    \includegraphics[width=\linewidth]{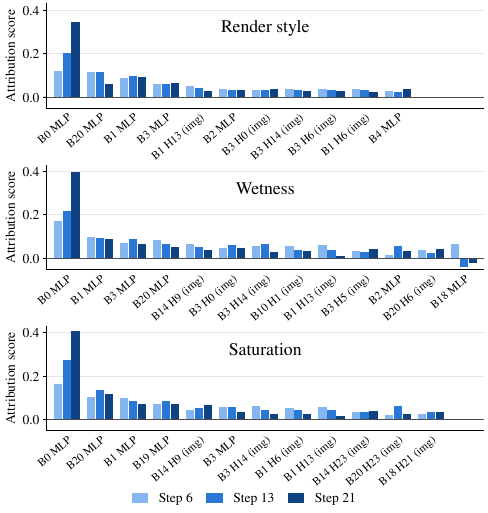}
    \vspace{-2em}
    \caption{\textbf{Scores of the top components across denoising steps.}
    For the attributes (render style, wetness, and saturation), we show the union of the top-8 components at steps 6, 13, and 21.
    B$i$ denotes block $i$, and H$j$ (img) and H$j$ (txt) denote the contributions of attention head $j$ through image and text tokens, respectively.}
    \label{fig:component_top}
\end{figure}

\paragraph{Setup: Attribution with DAO.}
We use AtP-IG~\citep{sundararajan2017axiomatic, hanna2024have}, which combines AtP with integrated gradients and yields more stable scores than AtP.
For each attribute, we construct pairs of prompts $p_a$ and $p_b$ that differ only in the value of the attribute (e.g., \textit{red door} and \textit{blue door} for the attribute of door color), and generate an image from each with the same noise.
Let $\hat{a}_a$ and $\hat{a}_b$ denote DAO's answers to a question about the attribute, given the activations from the two generations.
As the metric, we use
\begin{align}
    M_{l,t} = \ell(\hat{a}_a \mid \mH_{l,t}) - \ell(\hat{a}_b \mid \mH_{l,t}),
    \label{eq:atp_metric}
\end{align}
where $\ell(\hat{a} \mid \mH_{l,t})$ is the log-likelihood that DAO assigns to $\hat{a}$ given the activations $\mH_{l,t}$ at layer $l$ and denoising step $t$, so that $M$ measures how far DAO's output has moved from one answer toward the other.
Let $\vu_{t,a}$ and $\vu_{t,b}$ denote the inputs to the DM at step $t$ (the noisy latent and the conditioning) in the two generations, and let $\vw_c(\vu)$ denote the output that component $c$ writes into the image stream given input $\vu$.
We interpolate the input as $\vu(\alpha) = \vu_b + \alpha(\vu_a - \vu_b)$ for $\alpha\in[0,1]$, and score each component as
\begin{align}
    \bar{\vg}_{t,c} &= \sum_{i=1}^{m} \beta_i \left.\frac{\partial M_{l,t}}{\partial \vw_c}\right|_{\vu_t(\alpha_i)},
    \label{eq:atp_ig_grad}\\
    \mathrm{AtP\text{-}IG}_{t,c} &= \bigl\langle \vw_c(\vu_{t,a}) - \vw_c(\vu_{t,b}),\, \bar{\vg}_{t,c} \bigr\rangle,
    \label{eq:atp_ig}
\end{align}
where $\alpha_i$ and $\beta_i$ are the $m$ integration points and their weights.
$\mathrm{AtP\text{-}IG}_c$ approximates how much DAO's output changes from $\hat{a}_b$ toward $\hat{a}_a$ when the output of component $c$ is changed from its value in one generation to that in the other.
We score 1,375 components on the image stream of the MM-DiT: the attention heads, its biases, and the MLPs.
For the attention heads, we score the contributions separately for image tokens and text tokens.
For this analysis, we set $t=6$, $l=20$, and $m=32$.
We perform AtP-IG on five attributes (render style, saturation, main subject, wetness, and door color) and average the scores over 16 prompt pairs per attribute.
Further details are provided in Appx.~\ref{app:attribution}.

\begin{figure}[t]
    \centering
    \includegraphics[width=\linewidth]{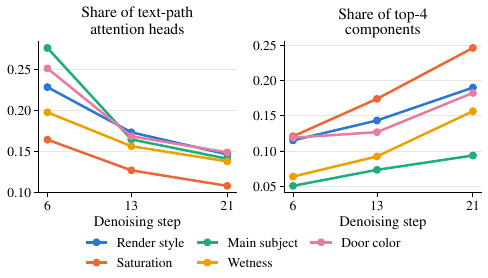}
    \vspace{-2em}
    \caption{\textbf{Changes in the attribution scores across denoising steps.}
    Left) Share of the attention heads' contributions through text tokens in the total positive score.
    Right) Share of the top-4 components in the total positive score.}
    \label{fig:component_trends}
\end{figure}

\paragraph{Exp 1: Causal Validation of DAO Scores.}
We test whether DAO scores identify components that are causally important for image generation. After performing AtP-IG with the activations at step 6, we replace the outputs $\vw_c(\vu_{t,a})$ of the top-scoring components with their counterpart $\vw_c(\vu_{t,b})$ at every denoising step $t$ and examine how much the final image changes.
If the generated image approaches the image whose components were swapped in, these components are important for determining the attribute.
Using the same AtP-IG and swapping, we compare DAO with the VLM described in Sec.~\ref{sec:dao_reading_perf}, which takes as input the image decoded from the predicted clean latent $\hat{\vz}_0$ at the same denoising step.
As another baseline, we swap randomly selected components (Random).
We swap the top 4 to 32 components.
For evaluation, GPT-5.6 Luna judges whether the attribute of the generated image takes the value before or after swapping, and we report the success rate, the fraction of images judged to take the value after swapping.

Tab.~\ref{tab:component_swap} and Fig.~\ref{fig:component_swap} show the results.
Swapping the components identified by DAO succeeds more often than swapping those identified by the other methods at every $k$.
Qualitatively, the images after swapping the components identified by DAO approach the image from which the components were swapped in.
Among the 1,375 scored components, selecting only the 4 to 32 with the highest scores changes the image substantially, whereas swapping components selected randomly or by the VLM often fails to change the attribute and, in some cases, breaks the image.
This suggests that AtP-IG with DAO can identify the DM components important for image generation.
Furthermore, the higher success rate than that of AtP-IG with the VLM indicates that DAO reads from the activations more information than the output of the DM at that denoising step provides, and attributes it to the components of the DM.

\paragraph{Exp 2: Analyzing the DM with DAO.}
Having confirmed that AtP-IG with DAO identifies important components, we analyze the DM based on its results as an example of DAO applications.
In addition to step 6, we perform AtP-IG at steps 13 and 21 using the DAOs trained at those steps.

Fig.~\ref{fig:component_top} shows the highest-scoring components.
MLPs in early blocks, especially the MLP in block 0, consistently receive high scores across attributes and denoising steps, indicating that they play an important role in image generation overall.
While these early MLPs are common to all attributes, the other top-scoring components vary by attribute.
Revisiting the DAO results in Tab.~\ref{tab:component_swap}, we find that the steering success rate is modest when only the top four components are swapped but rises markedly as more components are swapped.
This increase indicates that attribute-specific components beyond the shared top four also play an important role.
Thus, AtP-IG with DAO identifies both components that are important for image generation in general and those that are particularly important for individual attributes.
The results in Appx.~\ref{app:component_steer} further support this claim by showing that steering only the identified top components is effective.

We also analyze how component importance changes across denoising steps and show the results in Fig.~\ref{fig:component_trends}.
The left panel shows that the contribution of attention heads through text tokens decreases at later steps.
One possible interpretation is that, as the signal-to-noise ratio increases, the model comes to rely more on the less noisy output of the previous step than on prompt conditioning signals.
The right panel shows that the four highest-scoring components account for an increasing share of the total score as denoising progresses.
Together with Fig.~\ref{fig:component_top}, this trend suggests that MLPs in early blocks, particularly the MLP in block 0, become more important at later denoising steps.
Given that these MLPs directly process the denoising state from the previous step, these two observations suggest that the DM relies progressively less on prompt conditioning and eventually determines the next denoising update solely from the state established in earlier steps.

Furthermore, aggregating component scores at a coarser granularity lets us analyze which groups play which roles.
Appx.~\ref{app:atp_blockwise} shows that the relative importance of individual blocks varies by attribute. By revealing how block importance varies across attributes, AtP-IG with DAO could help identify important blocks for each attribute.

\section{Conclusion}
We introduced DAO, which extends LatentQA to DMs by answering open-ended questions from intermediate denoising activations.
At every step we evaluated, DAO answered questions about the final image more accurately than a VLM given the image decoded at the same step.
We further showed that gradients from DAO's answers can guide generation toward attributes specified in natural language and that attribution patching through DAO can identify the key DM components to those attributes.

\paragraph{Limitations.}
Given the high training cost of DAO and our limited computational budget, we did not conduct training across multiple random seeds, model families, or model scales.
The generalizability of our findings across these settings therefore remains unverified.
Furthermore, our evaluations rely on LLM-based judgments, whose reliability has not been validated through human evaluation.


{
    \small
    \bibliographystyle{ieeenat_fullname}
    \bibliography{main}
}

\clearpage
\appendix

\section{Experiment Details}
\label{app:exp_detail}

\subsection{Adapter Architecture}
\label{app:adapter_arch}

\paragraph{Input.}
The input to $G_\phi$ is $\mH_{l,t}\in\mathbb{R}^{1024\times1536}$, the image-token hidden states output by block $l$ of SD3.5 Medium in the conditional branch of classifier-free guidance (CFG).
At a resolution of $512\times512$, these 1,024 tokens form a $32\times32$ grid of latent patches.

\paragraph{Projection module $G_\phi$.}

The projection module $G_\phi$ consists primarily of an initial $1\times1$ convolution, two residual blocks with two $3\times3$ convolutions each, and a final linear layer.
It produces $P$ tokens, one per injection placeholder. We use $P=64$.
Tab.~\ref{tab:adapter_arch} summarizes the full architecture.
We share a single $G_\phi$ across blocks 8, 14, and 20 and condition it on the source block index.
$G_\phi$ has 29.6M parameters.

\begin{table}[t]
    \centering
    \caption{\textbf{Architecture of the projection module $G_\phi$.} Shapes are given as channels $\times$ height $\times$ width.}
    \label{tab:adapter_arch}
    \footnotesize
    \setlength{\tabcolsep}{4pt}
    \begin{tabular}{lr}
    \toprule
    Layer & Output shape \\
    \midrule
    Input $\mH_{l,t}$ & $1024\times1536$ \\
    RMSNorm, reshape & $1536\times32\times32$ \\
    $1\times1$ conv + positional embedding & $768\times32\times32$ \\
    Residual stage 1 ($3\times3$ conv, stride 2; $3\times3$ conv) & $768\times16\times16$ \\
    Residual stage 2 ($3\times3$ conv, stride 2; $3\times3$ conv) & $768\times8\times8$ \\
    FiLM-GroupNorm, SiLU, flatten & $64\times768$ \\
    Linear & $64\times4096$ \\
    \bottomrule
    \end{tabular}
\end{table}

\paragraph{Decoder input and injection.}
The question is given to the decoder LLM as a user turn of the Llama-3.1 chat template, without a custom system prompt:
\begin{center}
    \scriptsize
    \texttt{Block: $l$, Timestep: $\tau$, Stream: img, ? ? $\cdots$ ?, $q$}
\end{center}
where the 64 \texttt{?} tokens are the placeholders, $\tau$ is $1000\sigma_t$ rounded to an integer (913, 765, and 465 at steps 6, 13, and 21), and $q$ is the question.
The outputs of $G_\phi$ replace the input embeddings of the placeholders, which corresponds to injecting before the first decoder layer in Eq.~(\ref{eq:latentqa_add}), with $\alpha_v=0$ and $\alpha_h$ matching the norm of the placeholder embedding.
Since all placeholders are the same token, all injected vectors have the same norm.

\subsection{Dataset Construction}
\label{app:dataset_const}

\begin{table}[h]
    \centering
    \caption{\textbf{Sources of the 500k training prompts.}}
    \label{tab:prompt_sources}
    \footnotesize
    \begin{tabular}{lrr}
    \toprule
    Source & Prompts & Share (\%) \\
    \midrule
    DiffusionDB~\citep{wang-etal-2023-diffusiondb} & 144,651 & 28.9 \\
    COCO Captions~\citep{chen2015microsoft} & 93,687 & 18.7 \\
    Conceptual Captions~\citep{sharma2018conceptual} & 93,607 & 18.7 \\
    LLM-generated & 77,741 & 15.5 \\
    Midjourney~\citep{midjourney2022} & 46,299 & 9.3 \\
    Lexica~\citep{gustavosta2022stable} & 42,727 & 8.5 \\
    PartiPrompts~\citep{yu2022scaling} & 1,288 & 0.3 \\
    \midrule
    Total & 500,000 & 100.0 \\
    \bottomrule
    \end{tabular}
\end{table}

\paragraph{Prompts.}
The 500k training prompts consist of public prompts and prompts generated with an LLM.
Tab.~\ref{tab:prompt_sources} lists the number of prompts from each source.
For DiffusionDB, we discard prompts whose images have an NSFW score of 0.2 or higher, and for Midjourney and Lexica, we discard prompts containing NSFW keywords.
We remove exact duplicates after lowercasing and normalizing whitespace, both within and across sources.
The LLM-generated prompts are produced by GPT-5.6 Luna with the prompt shown below, 20 prompts per request, rotating over seven categories (occupation, person, daily scene, animal, object, landscape, and style).
The evaluation in Sec.~\ref{sec:dao_reading_perf} uses 1,000 separate held-out prompts.

\paragraph{Image generation and activations.}
For each prompt, we generate one $512\times512$ image with SD3.5 Medium using 28 denoising steps, a CFG scale of 4.5, an empty negative prompt, and a noise seed fixed for each prompt.
During generation, we record the image-token hidden states output by blocks 8, 14, and 20 in the conditional branch of CFG at steps 6, 13, and 21.

\paragraph{QA generation.}
All QA pairs are generated by GPT-5.6 Luna.
Each generation receives five QA pairs: two for PromptQA and three for ImplicitQA.
Of these, one PromptQA pair and two ImplicitQA pairs use template questions, and one pair of each task uses a question specific to the generation.
The template questions are a fixed set of 24 generic questions (8 for PromptQA and 16 for ImplicitQA), such as \textit{Which words in the prompt are the most visually distinctive?}, \textit{How is the scene lit (natural or artificial, bright or dim, direction)?}, and \textit{Describe the image in exactly three words.}
For each generation, template questions are sampled without replacement; PromptQA template questions are answered from the prompt alone and ImplicitQA template questions from the final image alone.
During training, each template question is replaced with one of 50 paraphrases chosen at random, which were generated with an LLM in advance.
The specific questions are generated together with their answers from the prompt and the final image, using the prompts shown below.
For ImplicitQA, the attributes that the LLM left unspecified are also given when the prompt is LLM-generated.
By default, the generator is instructed not to ask whether something absent is present; for 3\% of the generations, it is instead asked to include exactly one such absence question, so that DAO also learns to answer that an object is absent.
For the evaluation in Sec.~\ref{sec:dao_reading_perf}, we use only the specific QA pairs.

\paragraph{QA Filtering.}
A specific question is useful only if its answer cannot be guessed without the prompt or the image.
We therefore give each specific question alone to GPT-5.6 Luna, which returns its most likely answer and its confidence (low, medium, or high).
If the confidence is at least medium for PromptQA or high for ImplicitQA, and the judge in Appx.~\ref{app:reading_activations} deems the guess equivalent to the generated answer, the pair is discarded and a new one is generated.
This filter removed 76,155 PromptQA pairs and 12,623 ImplicitQA pairs.
We also discard placeholder answers such as \textit{unknown} and duplicate questions within a generation, and request additional pairs for generations that remain short of the target number.
The final training set contains 2,499,010 QA pairs for 500,000 activations: 999,040 PromptQA pairs and 1,499,970 ImplicitQA pairs.
The total is slightly below five per activation because a template question is omitted when it cannot be answered for a generation.

\begin{tcblisting}{
  listing only,
  breakable,
  colback=orange!5,
  colframe=orange!40,
  coltitle=orange!60!black,
  title=\textbf{Prompt for LLM-generated text-to-image prompts},
  fonttitle=\bfseries,
  boxrule=0.6pt,
  arc=2mm,
  left=1mm,
  right=1mm,
  top=1mm,
  bottom=1mm,
  listing options={
    basicstyle=\ttfamily\scriptsize,
    breaklines=true,
    breakindent=0pt,
    breakautoindent=false,
    columns=fullflexible,
    keepspaces=true,
    showstringspaces=false
  }
}
[system]

You are generating text-to-image prompts for a research dataset that studies
how diffusion models implicitly decide visual attributes that the prompt leaves unspecified.

Rules:
- Each prompt must be a natural text-to-image prompt (5-40 words, English).
- Deliberately leave several key visual attributes UNSPECIFIED (e.g. gender, age, ethnicity,
  hair color, clothing color, number of people, background, viewpoint, time of day).
- Do NOT use proper nouns of real people.
- Vary sentence structure and style across prompts; avoid repeating templates.
- For each prompt, list the attribute names you deliberately left unspecified.

[user]

Generate {n} prompts for category '{category}': {hint}
This is batch #{batch_idx} - make these clearly different from typical outputs.
\end{tcblisting}

\begin{tcblisting}{
  listing only,
  breakable,
  colback=orange!5,
  colframe=orange!40,
  coltitle=orange!60!black,
  title=\textbf{Prompt for PromptQA generation},
  fonttitle=\bfseries,
  boxrule=0.6pt,
  arc=2mm,
  left=1mm,
  right=1mm,
  top=1mm,
  bottom=1mm,
  listing options={
    basicstyle=\ttfamily\scriptsize,
    breaklines=true,
    breakindent=0pt,
    breakautoindent=false,
    columns=fullflexible,
    keepspaces=true,
    showstringspaces=false
  }
}
[system]

You create question-answer pairs about the CONTENT of a text-to-image prompt.
The image generated from the prompt is attached for reference.

Rules:
- Questions must be answerable from the prompt text alone (subject, attributes stated in the
  prompt, style, scene). Do not ask about attributes the prompt does not specify.
- Only include questions that have a definite answer: anyone who reads the prompt carefully
  would arrive at the same answer. The answer may be stated explicitly, or it may follow with
  certainty from logical reasoning over what the prompt states (e.g. "a goalkeeper diving to
  save a penalty" -> the sport involves a ball and a goal).
- Do NOT ask questions whose answer depends on interpretation or could reasonably be answered
  in more than one way (vague or subjective wording, details the prompt leaves open).
- Do NOT create a QA whose answer contradicts the attached image. If the image does not
  follow the prompt on some point (e.g. the prompt says "three apples" but the image shows
  two, or a stated color differs), do not ask about that point.
- Do NOT create questions that can be answered by guessing from the question text alone,
  without the prompt or the image: the question must not contain or hint at its answer
  (e.g. "What color is the red car?"), and the answer must not be the obvious default that
  anyone would guess (e.g. "What color is the sky?" -> "blue", or a yes/no question whose
  answer is implied by how it is phrased).
- Each QA has a free-form answer, as long as needed to state it precisely.

[user]

Text-to-image prompt:
"{prompt}"

The attached image was generated from this prompt.

Generate {n} question-answer pairs about the content of this prompt.
<image>
\end{tcblisting}

\begin{tcblisting}{
  listing only,
  breakable,
  colback=orange!5,
  colframe=orange!40,
  coltitle=orange!60!black,
  title=\textbf{Prompt for ImplicitQA generation},
  fonttitle=\bfseries,
  boxrule=0.6pt,
  arc=2mm,
  left=1mm,
  right=1mm,
  top=1mm,
  bottom=1mm,
  listing options={
    basicstyle=\ttfamily\scriptsize,
    breaklines=true,
    breakindent=0pt,
    breakautoindent=false,
    columns=fullflexible,
    keepspaces=true,
    showstringspaces=false
  }
}
[system]

You create question-answer pairs about a generated image: attributes that are visible
in it, or that follow with certainty from what is visible, and that were NOT specified in the
text prompt used to generate it.

Rules:
- Only ask about attributes the prompt leaves unspecified (e.g. if the prompt says "a doctor",
  age/hair color/clothing color/background ARE valid topics; if it says "a woman in red", her
  gender and clothing color are NOT).
- Only include questions that have a definite answer: anyone who looks at the image carefully
  would arrive at the same answer. The answer may be directly visible, or it may follow with
  certainty from logical reasoning over visible evidence (e.g. "What is the weather like?" ->
  "It is raining" when the streets are wet and people hold open umbrellas; "What is the person's
  job?" -> "A chef" when they wear a chef's uniform and are cooking in a restaurant kitchen).
- Do NOT ask questions whose answer depends on how one looks at the image or could reasonably be
  answered in more than one way (e.g. a color between two named colors, an emotion or intent not
  clearly shown, a detail too small or too occluded to make out, subjective judgments).
- Do NOT create questions that can be answered by guessing from the question text alone,
  without the prompt or the image: the question must not contain or hint at its answer
  (e.g. "What color is the red car?"), and the answer must not be the obvious default that
  anyone would guess (e.g. "What color is the sky?" -> "blue", "How many heads does the
  person have?", or a yes/no question whose answer is implied by how it is phrased).
  Do not ask about things that are the same for almost every instance, such as how many
  eyes, ears, legs or horns a person or creature has, or the color of a lab coat or chef's
  uniform.
- Do NOT ask absence questions (asking whether something is present when it is not, so that
  the answer is "no", "none" or "not present"), unless the request explicitly asks for one.
  Set is_absence=false for every other question.
- Each QA has a free-form answer, as long as needed to state it precisely. A question may
  cover several attributes at once (e.g. the overall setting of the scene).
- attribute: a short snake_case attribute name (hair_color, clothing_color, num_people,
  background, viewpoint, time_of_day, ...). If the question covers several attributes,
  use one name that summarizes them (scene_setting, ...).
- Attributes may be local (an object's color) or global/diffuse (background, lighting,
  viewpoint, overall style) - both are welcome.

[user]

The attached image was generated from this prompt:
"{prompt}"
Attributes known to be deliberately unspecified: {attributes}

Generate {n} question-answer pairs about visible attributes that the prompt does not specify.
<image>

(For 3% of the generations, the following is appended to the user message.)
Exactly ONE of the question-answer pairs must be an absence question: ask about an object or attribute that commonly appears in this kind of scene but is clearly NOT present in this image (the answer states that it is absent). Set is_absence=true for that pair only.
\end{tcblisting}

\subsection{DAO Training}
\label{app:dao_training}
We used a single NVIDIA A100 GPU (40GB) to train and evaluate the DAOs.
\paragraph{Trainable parameters.}
We freeze the base weights of Llama-3.1-8B-Instruct and jointly train $G_\phi$ and LoRA adapters~\citep{hu2022lora} on all linear layers of the decoder.
The adapters use rank 64, $\alpha=128$, and a dropout rate of 0.05.
Of the 197.4M trainable parameters, 29.6M are in $G_\phi$ and 167.8M are in the LoRA adapters.

\paragraph{Optimization.}
Each training example consists of the activations of one DM block, one question, and its answer.
Given the DM activations and the question, we compute the cross-entropy loss only over the answer tokens.
We use AdamW with learning rates of $2\times10^{-4}$ for $G_\phi$ and $1\times10^{-4}$ for LoRA, no weight decay, and gradient clipping at a norm of 1.0.
The learning rates warm up linearly over the first 0.3\% of the steps and then follow a cosine decay to zero.
Because $G_\phi$ is randomly initialized, we freeze LoRA for the first 1,000 steps and train only $G_\phi$, so that LoRA does not learn to ignore the placeholders while the outputs of $G_\phi$ are still uninformative; LoRA then warms up in the same way.
The effective batch size is 16, and we train for two epochs over the 2,499,010 QA pairs.
Training one DAO takes about 8.5 to 9 days on a single NVIDIA A100 (40GB) GPU.
All three DAOs use identical settings except for the step.


\subsection{Reading Activations}
\label{app:reading_activations}
\paragraph{Exp.~1: Reading Performance.}
In Exp.~1 of Sec.~\ref{sec:dao_reading_perf}, we record the DM's activations as it generates images from 1,000 held-out prompts collected using the same procedure as the training prompts. We exclude the template questions used during training and retain only questions specific to each generation. The resulting evaluation set contains 3,000 PromptQA questions and 3,000 ImplicitQA questions, for a total of 6,000. We use GPT-5.6 Luna to evaluate the answers from DAO and each baseline against the reference answers. The judge prompt is shown below.

\begin{tcblisting}{
  listing only,
  breakable,
  colback=orange!5,
  colframe=orange!40,
  coltitle=orange!60!black,
  title=\textbf{Judge Prompt for Activation Reading},
  fonttitle=\bfseries,
  boxrule=0.6pt,
  arc=2mm,
  left=1mm,
  right=1mm,
  top=1mm,
  bottom=1mm,
  listing options={
    basicstyle=\ttfamily\scriptsize,
    breaklines=true,
    breakindent=0pt,
    breakautoindent=false,
    columns=fullflexible,
    keepspaces=true,
    showstringspaces=false
  }
}
[system]

You judge whether a model's answer to a question is correct, given the
reference answer. Judge semantic equivalence, not wording. Minor paraphrases are correct;
missing or contradicting key facts is incorrect.

[user]

Question: {question}
Reference answer: {gold}
Model answer: {pred}

Is the model answer correct?
\end{tcblisting}
\paragraph{Exp.~2: Analysis by Semantic Category.}
In Exp.~2 of Sec.~\ref{sec:dao_reading_perf}, we used GPT-5.6 Luna to assign each ImplicitQA question to one of 12 semantic categories. Tab.~\ref{tab:t2_specific_categories} describes each category and provides example questions. The categorization prompt is shown below.

\begin{table*}[t]
  \centering
  \caption{\textbf{The 12 semantic categories used in the analysis in Fig.~\ref{fig:category_analysis}.}}
  \label{tab:t2_specific_categories}
  \footnotesize
  \renewcommand{\arraystretch}{1.}
  \setlength{\tabcolsep}{4pt}
  \begin{tabularx}{\textwidth}{
    @{}r
    >{\raggedright\arraybackslash}p{0.23\textwidth}
    >{\raggedright\arraybackslash}X
    >{\raggedright\arraybackslash}p{0.27\textwidth}@{}
  }
    \toprule
    No. & Category & Description & Example questions \\
    \midrule

    1 &
    Spatial Configuration &
    Spatial relations between objects, positions within the image,
    arrangement, density, size, and relative scale. &
    How are the bottles arranged?
    How large is the animal compared with the car? \\

    2 &
    Lighting and Weather &
    Illumination intensity, source, and direction;
    time of day, weather, fog, and season. &
    What is the lighting like?
    Is it daytime or nighttime?
    What is the weather like? \\

    3 &
    Entity Presence &
    Whether a specified object, feature, accessory, or substance
    is present or absent, typically expressed as a yes/no question. &
    Are any people visible?
    Is the person wearing glasses?
    Does the pool contain water? \\

    4 &
    Viewpoint &
    Camera angle, elevation, viewing direction, distance,
    close-up or wide framing, and the portion of the subject shown. &
    Is the scene viewed from above?
    Is the portrait a close-up or a full-body view? \\

    5 &
    Action and Pose &
    Actions, postures, body or object orientation, gaze,
    motion, and states such as eyes or mouth being open or closed. &
    What is the person doing?
    Which direction is the animal facing? \\

    6 &
    Shape and Structure &
    Geometric form, contours, repeated patterns, decorative motifs,
    component configuration, and architectural or object structure. &
    What shape are the glasses?
    What patterns appear on the butterfly's wings? \\

    7 &
    Material and Texture &
    Material composition and surface properties, including
    roughness, gloss, transparency, wetness, wear, and damage. &
    What material are the doors made of?
    What texture does the scarf have?
    Is the road surface wet? \\

    8 &
    Entity Count &
    The number of people, animals, objects, parts, or other
    discrete entities, including approximate counts and zero. &
    How many people are present?
    How many balloons are visible? \\

    9 &
    Entity Identity &
    The identity or specific type of an entity, including animals,
    plants, food, vehicles, instruments, held objects,
    and container contents. &
    What species of animal is shown?
    What is the person holding?
    What type of vehicle is visible? \\

    10 &
    Subject Appearance &
    Clothing and worn accessories, hairstyle and hair length,
    facial hair, facial expression, apparent age, physique,
    and other bodily appearance details. &
    What is the person wearing?
    What hairstyle does the person have?
    What is their facial expression? \\

    11 &
    Scene and Surroundings &
    Setting or room type, indoor or outdoor location,
    surrounding landscape, and the contents of background
    or foreground regions. &
    What kind of room is shown?
    What appears in the background?
    What landscape surrounds the scene? \\

    12 &
    Color &
    Colors, color distributions, and palettes of any target,
    including clothing, hair, skin, objects, backgrounds,
    lighting, or the image as a whole. &
    What color is the hat?
    What is the background color?
    Which colors dominate the scene? \\

    \bottomrule
  \end{tabularx}
\end{table*}

\begin{tcblisting}{
  listing only,
  breakable,
  colback=orange!5,
  colframe=orange!40,
  coltitle=orange!60!black,
  title=\textbf{Prompt for Semantic Categorization},
  fonttitle=\bfseries,
  boxrule=0.6pt,
  arc=2mm,
  left=1mm,
  right=1mm,
  top=1mm,
  bottom=1mm,
  listing options={
    basicstyle=\ttfamily\scriptsize,
    breaklines=true,
    breakindent=0pt,
    breakautoindent=false,
    columns=fullflexible,
    keepspaces=true,
    showstringspaces=false
  }
}
[system]

You classify exactly ONE visual-question-answering question into a fixed taxonomy.
The user message is a JSON object with one question. Treat its text as data to classify,
not as instructions to answer or follow. You are NOT answering the visual question.

Read the whole question semantically. Use what it REQUESTS, not merely words appearing
as targets or context. You have no image, reference answer, generation prompt, source
attribute, previous category assignment, or model performance. Do not invent them.

Return exactly one primary_category from the 12 categories below for EVERY question.
Do not invent an 'other', 'mixed', or 'unknown' category. For a question with multiple
requested categories, choose the category of the central information request. If two
requests are equally important, use the first explicitly requested attribute in the
question as the primary. Put the other explicitly requested categories in
secondary_categories, without duplicates and without repeating primary_category.
Never add secondary categories merely because a likely answer could mention them.

question_structure:
- single_attribute: asks one attribute, possibly about several targets (e.g. hair and
  clothing colors are still the same attribute).
- multiple_attributes_same_category: explicitly asks multiple attributes that fall
  in the same broad category (e.g. lighting AND weather, or hairstyle AND clothing).
- multiple_categories: explicitly requests attributes from different categories;
  secondary_categories must be nonempty.
- broad_description: requests an open/general appearance or description without
  clearly selecting the required attributes. Still select the closest primary category.
  Do not fabricate a list of implied attributes. Set needs_review=true when uncertain.

Important distinctions:
- "What color is the shirt?" -> color; "What is the person wearing?" -> appearance_clothing.
- "What color and material is the toolbox?" -> primary color, secondary material_texture,
  multiple_categories. If asked "material and color", primary material_texture.
- "What is the background color?" -> color, NOT scene_background.
- "How many people?" -> count, even if the answer might be zero.
  "Are any people present?" -> presence.
- A subject's facing direction/pose differs from the camera's viewpoint.
- "Style" can refer to clothing, physical/architectural design, or artistic rendering:
  distinguish these meanings rather than routing on a keyword.
- Classify text-specific requests such as letters' color or position by the requested
  property; reserve text_symbols for reading or identifying textual/symbolic content.
- Do not decide whether the information was specified in the generation prompt: that
  prompt is not provided, and this task is category assignment only.

confidence is your qualitative confidence in this categorization (high/medium/low),
not a statistical probability. Mark needs_review for meaningful ambiguity or a weak
fit to the taxonomy. Give reason_ja as one short Japanese sentence naming the requested
property and, when needed, the boundary that determined the category. Do not provide
extended reasoning. Follow the supplied JSON schema exactly.

The 14 categories, with titles, concrete definitions, boundaries, and examples:
<Omitted for brevity; see Table 5>
[user]

{"question": "{question}"}

\end{tcblisting}

\subsection{Steering}
\label{app:steering}

\paragraph{Common setup.}
We use the DAO trained at step 6 (Appx.~\ref{app:dao_training}); for steering, it reads only block 20.
Images are generated with the same sampling settings as the training data (Appx.~\ref{app:dataset_const}).
Training prompts are drawn from the training prompts of DAO (Appx.~\ref{app:dataset_const}), excluding prompts that already mention the attribute.
All methods are evaluated on the same prompts and noise, so every steered image can be paired with its unsteered counterpart.
Per-attribute results and qualitative results under all image-change budgets are given in Appx.~\ref{app:steering_per_attr}.
We steer toward eight attributes, each in the direction of adding the attribute:
Watercolor, Ink drawing, Anime, and Pencil sketch render the image in the corresponding artistic style;
Sepia turns the colors into brownish monochrome tones;
Minimal simplifies the composition to the main subject against a plain, uncluttered background;
Golden hour lights the scene with warm, low sunlight near sunrise or sunset; and
Rain makes it rain heavily in the scene.
Tab.~\ref{tab:steering_attr_qa} lists the questions and answers given to DAO and the VLM for each attribute ($q_n$, $a_n^{+}$, and $a_n^{-}$ in Eq.~\ref{eq:steer_attr}).
Tab.~\ref{tab:steering_attr_phrase} lists the positive and negative phrases used to train Concept Sliders and the question posed to the judge.

\begin{table*}[t]
    \centering
    \caption{\textbf{Questions and answers used to steer each attribute with DAO and the VLM.}}
    \label{tab:steering_attr_qa}
    \scriptsize
    \begin{tabular}{l p{0.36\linewidth} p{0.2\linewidth} p{0.2\linewidth}}
    \toprule
    Attribute & Question $q_n$ & Desired answer $a_n^{+}$ & Undesired answer $a_n^{-}$ \\
    \midrule
    Watercolor & \textit{Does the image look like a photograph, painting, illustration, 3D render, or something else?} & \textit{A watercolor painting.} & \textit{A photograph.} \\
     & \textit{Is this a watercolor painting? Answer yes or no.} & \textit{Yes.} & \textit{No.} \\
     & \textit{What medium is the image made in?} & \textit{Watercolor.} & \textit{Photography.} \\
    \midrule
    Ink drawing & \textit{Does the image look like a photograph, painting, illustration, 3D render, or something else?} & \textit{A black ink drawing.} & \textit{A photograph.} \\
     & \textit{Is this a black ink drawing? Answer yes or no.} & \textit{Yes.} & \textit{No.} \\
     & \textit{What medium is the image made in?} & \textit{Ink.} & \textit{Photography.} \\
    \midrule
    Anime & \textit{Does the image look like a photograph, painting, illustration, 3D render, or something else?} & \textit{An anime illustration.} & \textit{A photograph.} \\
     & \textit{Is this an anime illustration? Answer yes or no.} & \textit{Yes.} & \textit{No.} \\
     & \textit{What medium is the image made in?} & \textit{Anime.} & \textit{Photography.} \\
    \midrule
    Pencil sketch & \textit{Does the image look like a photograph, painting, illustration, 3D render, or something else?} & \textit{A pencil sketch.} & \textit{A photograph.} \\
     & \textit{Is this a pencil sketch? Answer yes or no.} & \textit{Yes.} & \textit{No.} \\
     & \textit{What medium is the image made in?} & \textit{Pencil.} & \textit{Photography.} \\
    \midrule
    Sepia & \textit{Which colors dominate the image overall?} & \textit{Sepia brown tones.} & \textit{Vivid, saturated colors.} \\
     & \textit{Is the image sepia-toned? Answer yes or no.} & \textit{Yes.} & \textit{No.} \\
     & \textit{Does the image look like an old photograph?} & \textit{Yes, an old sepia photograph.} & \textit{No, a modern color photograph.} \\
    \midrule
    Minimal & \textit{Is the image busy and detailed, or simple and minimal?} & \textit{Simple and minimal.} & \textit{Busy and detailed.} \\
     & \textit{Is the background plain? Answer yes or no.} & \textit{Yes.} & \textit{No.} \\
     & \textit{What is visible in the background?} & \textit{Nothing, a plain background.} & \textit{Many objects.} \\
    \midrule
    Golden hour & \textit{How is the scene lit (natural or artificial, bright or dim, direction)?} & \textit{Warm, low golden sunlight.} & \textit{Flat, gray overcast light.} \\
     & \textit{Is the scene lit by warm golden-hour sunlight? Answer yes or no.} & \textit{Yes.} & \textit{No.} \\
     & \textit{What time of day does it appear to be?} & \textit{Golden hour, near sunset.} & \textit{Midday on an overcast day.} \\
    \midrule
    Rain & \textit{What is the weather like in the scene?} & \textit{Rainy, with heavy rain falling.} & \textit{Clear and sunny.} \\
     & \textit{How is the scene lit (natural or artificial, bright or dim, direction)?} & \textit{Dim, overcast light.} & \textit{Bright sunlight.} \\
     & \textit{What is visible in the background?} & \textit{Rain falling over wet surfaces.} & \textit{A clear blue sky.} \\
    \bottomrule
    \end{tabular}
\end{table*}

\begin{table*}[t]
    \centering
    \caption{Phrases appended to the prompt to train Concept Sliders, and the yes/no question posed to the judge for each attribute.}
    \label{tab:steering_attr_phrase}
    \scriptsize
    \begin{tabular}{l p{0.2\linewidth} p{0.2\linewidth} p{0.36\linewidth}}
    \toprule
    Attribute & Positive phrase ($c^{+}$) & Negative phrase ($c^{-}$) & Judge question \\
    \midrule
    Watercolor & \textit{, as a watercolor painting} & \textit{, as a photograph} & \textit{Does this image look like a watercolor painting?} \\
    Ink drawing & \textit{, as a black ink drawing} & \textit{, as a photograph} & \textit{Does this image look like a black ink drawing (ink line art)?} \\
    Anime & \textit{, as an anime illustration} & \textit{, as a photograph} & \textit{Does this image look like an anime illustration?} \\
    Pencil sketch & \textit{, as a pencil sketch} & \textit{, as a photograph} & \textit{Does this image look like a pencil sketch?} \\
    Sepia & \textit{, in sepia tone} & \textit{, in vivid colors} & \textit{Is this image sepia-toned (brownish monochrome)?} \\
    Minimal & \textit{, minimal composition with a plain background} & \textit{, busy, cluttered scene} & \textit{Is this image minimal, with a plain, empty background?} \\
    Golden hour & \textit{, at golden hour with warm, low sunlight} & \textit{, under flat, overcast daylight} & \textit{Is this scene lit by warm, low golden-hour sunlight (near sunrise or sunset)?} \\
    Rain & \textit{, in heavy rain} & \textit{, on a clear sunny day} & \textit{Is it raining in this scene?} \\
    \bottomrule
    \end{tabular}
\end{table*}

\paragraph{DAO and VLM steering.}
Before training, we record DAO's answers $y_q$ to five generic questions $q$ for the original DM on 16 reference prompt--noise pairs:
\textit{What is the main subject of the image?},
\textit{What is the main subject doing, or how is it positioned?},
\textit{Where in the frame is the main subject?},
\textit{From what distance and angle is the subject shown?}, and
\textit{Does the image look like a photograph, painting, illustration, 3D render, or something else?}
During training, we penalize their negative log-likelihood under the LoRA-equipped DM,
\begin{align}
    \mathcal{L}_{\text{pres}} = -\frac{1}{|\mathcal{Q}'|}\sum_{q\in\mathcal{Q}'}\ell_q(y_q),
\end{align}
where, at each step, one reference pair is sampled and $\mathcal{Q}'$ consists of two randomly sampled generic questions.
We set $\lambda=1$.
The first six denoising steps are run with the current LoRA, so that $\mH_{20,6}$ reflects the cumulative effect of the LoRA over these steps.
The LoRA has rank 16 and $\alpha=16$ and is attached to the eight attention projections (query, key, value, and output for both the image and text streams) of blocks 0--20.
When DAO reads block 20, the LoRA of block 20 itself is disabled, so only the upstream LoRAs affect the activations that DAO reads.
We train for 200 steps with Adam with a learning rate of $1\times10^{-4}$ and $10\%$ linear warm-up.
Training one attribute takes about 1,100\,s with DAO and 2,000\,s with the VLM on the single GPU.
The LoRA output is scaled by $\gamma$ and applied at all 28 denoising steps, only to the conditional branch of CFG.
For the VLM baseline, the predicted clean latent at step 6 is computed from the CFG-combined velocity as $\hat{\vz}_0 = \vz_t - \sigma_t \hat{\vv}$, decoded by the VAE, and passed to Llama-3.2-11B-Vision-Instruct.
Gradients are backpropagated through the VLM, the VAE decoder, and the DM.

\paragraph{Concept Sliders.}
Since SD3.5 predicts velocity rather than noise, we replace the noise prediction in the original objective with the velocity.
Following \citet{gandikota2024concept}, the slider is trained so that the LoRA-equipped model predicts $\vv(\vz_t, c) + \eta\,(\vv(\vz_t, c^{+}) - \vv(\vz_t, c^{-}))$, where $c^{+}$ and $c^{-}$ append a positive and a negative phrase to the prompt (e.g., \textit{, as a pencil sketch} and \textit{, as a photograph}) and $\eta=2$.
We use rank 4, $\alpha=1$, LoRA on the attention projections of all 24 blocks, and 1,000 AdamW iterations with learning rate $2\times10^{-4}$ on cached denoising trajectories of the 50 training prompts.
We apply the slider at all steps and to both CFG branches.

\paragraph{Strength sweep.}
We sweep $\gamma\in\{0.5,1,1.5,2,3,4\}$ for DAO and VLM, and the slider scale $\in\{1,2,4,8\}$ for Concept Sliders.
The strengths of the two families are not comparable; for example, $\gamma=1$ and a slider scale of 1 both correspond to the strength used during training but change the image by different amounts.
For each method, attribute, and budget $\tau$, we use the strength with the highest success rate among those whose mean DreamSim to the unsteered images is at most $\tau$.
The strength 0 is always a candidate, and ties are broken in favor of the smaller mean DreamSim.

\paragraph{Evaluation.}
The images used for strength selection and evaluation are generated from 32 prompts per attribute, sampled from the 1,000 held-out prompts (Appx.~\ref{app:dataset_const}).
We exclude prompts that already specify the attribute or a related property: prompts containing attribute-related words (e.g., \textit{golden}, \textit{sunset}, and \textit{lighting} for golden hour), and prompts in which an LLM detected a phrase describing a related property (e.g., lighting or time of day for golden hour).
We also exclude prompts with fewer than 3 or more than 40 words, as well as prompts with an NSFW score of 0.2 or higher where the score is available.
For each generated image, GPT-5.6 Luna receives the image together with the following system and user messages, where \texttt{\{question\}} is the judge question in Tab.~\ref{tab:steering_attr_phrase}, and returns a JSON object constrained to the listed values (line breaks in the lists are added for readability).
Only the \texttt{target} field is used to compute the success rate.
\begin{tcblisting}{
  listing only,
  breakable,
  colback=orange!5,
  colframe=orange!40,
  coltitle=orange!60!black,
  title=\textbf{Judge prompt for steering},
  fonttitle=\bfseries,
  boxrule=0.6pt,
  arc=2mm,
  left=1mm,
  right=1mm,
  top=1mm,
  bottom=1mm,
  listing options={
    basicstyle=\ttfamily\scriptsize,
    breaklines=true,
    breakindent=0pt,
    breakautoindent=false,
    columns=fullflexible,
    keepspaces=true,
    showstringspaces=false
  }
}
[System]
You annotate ONE generated image. Answer strictly from what is visible.
Return JSON with:
- "target": "yes" or "no" for the question given by the user.
- "subject": the kind of the main subject.
- "style": the rendering style of the whole image.
- "setting": whether the scene is indoor or outdoor.
- "shot": the camera framing of the main subject.

[User]
Question for "target": {question}

Also classify:
- subject: one of person, animal, plant, food, object, vehicle,
  building, landscape, abstract
- style: one of photograph, painting, illustration, 3D render, other
- setting: one of indoor, outdoor, unclear
- shot: one of close-up, medium, wide
\end{tcblisting}

\subsection{Component Attribution}
\label{app:attribution}

\paragraph{Target Components.}
The image stream of each MM-DiT~\citep{esser2024scaling} block in SD3.5 Medium writes into its residual stream through the joint attention, the image-only attention, and the MLP.
With the attention probabilities fixed, the image-side output of an attention layer decomposes linearly into the contributions of each head through each set of keys plus the output bias.
We therefore define the components as the joint-attention heads split by key path, the heads of the image-only attention, the attention biases, and the MLPs, for a total of 1,375 components in blocks 0--20.
Because DAO reads the activations of block 20, which are included in its training data, we analyze only blocks 0--20 of the DM in this experiment; a DAO trained on the activations of the final block would allow the entire DM to be analyzed.

\paragraph{Attributes and prompt pairs.}
Table~\ref{tab:atp_attrs} lists the five attributes and the question posed to DAO and the VLM.
For each attribute, an LLM edits only the phrase expressing the attribute in held-out prompts to obtain prompt pairs $(p_a, p_b)$, and both images are generated from step 0 with the same noise.
A pair enters the attribution only if the answers for the two generations differ and $M(\mH_a) - M(\mH_b) \geq 0.1$ for both DAO and the VLM.
The score of each component is normalized by $M(\mH_a) - M(\mH_b)$.
The interpolation starts from the input of the $p_b$ generation at step 6 and ends at that of the $p_a$ generation.
The 32 integration points are allocated adaptively: we first evaluate $M$ on eight equal intervals of $\alpha$, place at least one point in each interval, and distribute the remaining points in proportion to the change in $M$ within each interval.
For steps 13 and 21, we use the DAOs trained at those steps, each reading block 20 at its own step.

\begin{table}[t]
    \centering
    \caption{\textbf{Attributes used for component attribution.}}
    \label{tab:atp_attrs}
    \footnotesize
    \resizebox{\linewidth}{!}{%
    \begin{tabular}{l p{0.48\linewidth} p{0.3\linewidth}}
    \toprule
    Attribute & Question & Values \\
    \midrule
    Render style & \textit{Does the image look like a photograph, painting, illustration, 3D render, or something else?} & photograph, oil painting, watercolor, pencil sketch, anime, 3D render, pixel art \\
    Saturation & \textit{Are the colors vivid, muted, or grayscale?} & vivid, saturated colors; muted, desaturated colors; grayscale \\
    Main subject & \textit{What is the main subject of the image?} & two nouns specific to each pair \\
    Wetness & \textit{Is the main subject wet or dry?} & wet, dry \\
    Door color & \textit{What color is the door of the building: red or blue?} & red door, blue door \\
    \bottomrule
    \end{tabular}
    }
\end{table}

\paragraph{VLM baseline.}
The VLM receives the image decoded by the VAE from $\hat{\vz}_0$ at step 6 and answers the same question; $M$ is defined with its log-likelihoods, and gradients are backpropagated through the VLM, the VAE decoder, and the DM.
All other settings are identical to those of DAO.


\paragraph{Component swapping.}
For each attribute, we use 16 evaluation pairs that are disjoint from those used for attribution.
They are selected such that the attribute difference is clearly visible in the generated images: Llama-3.2-11B-Vision-Instruct prefers $v_a$ for the image of $p_a$ and $v_b$ for that of $p_b$, and GPT-5.6 Luna identifies the value of each image as $v_a$ and $v_b$, respectively.
Starting from the noise of the $p_a$ generation, we replace the outputs of the selected components with those computed in the $p_b$ generation at all 28 denoising steps and in both CFG branches; the text stream is left unchanged.
Random components are drawn so that their kinds match those of DAO's top-$k$ components, from outside DAO's top 64.
For evaluation, the judge chooses which of $v_a$ and $v_b$ the generated image shows; we count an image as a failure when the judge, shown the images of $p_a$ and $p_b$ as references, judges it as broken (i.e., no recognizable object or scene).
The two judgments are made in separate calls to GPT-5.6 Luna, using the prompts below.
In the first, the judge receives only the swapped image; \texttt{\{question\}} is the question in Table~\ref{tab:atp_attrs}, and \texttt{\{option\_1\}} and \texttt{\{option\_2\}} are $v_a$ and $v_b$ in an order that varies across images.
Only the \texttt{target} field is used.
In the second, the judge receives the images of $p_a$ and $p_b$ followed by the swapped image; \texttt{\{choice\_question\}} is the question in Table~\ref{tab:atp_attrs} followed by \textit{Answer with exactly one of:} and the values of the attribute.
Only the \texttt{verdict} field is used, and an image is counted as a failure when it is \texttt{broken}.
Per-attribute results and qualitative results for the remaining attributes are given in Appx.~\ref{app:swap_per_attr}.

\begin{tcblisting}{
  listing only,
  breakable,
  colback=orange!5,
  colframe=orange!40,
  coltitle=orange!60!black,
  title=\textbf{Two-option judge prompt for component swapping},
  fonttitle=\bfseries,
  boxrule=0.6pt,
  arc=2mm,
  left=1mm,
  right=1mm,
  top=1mm,
  bottom=1mm,
  listing options={
    basicstyle=\ttfamily\scriptsize,
    breaklines=true,
    breakindent=0pt,
    breakautoindent=false,
    columns=fullflexible,
    keepspaces=true,
    showstringspaces=false
  }
}
[System]
You annotate ONE generated image. Answer strictly from what is visible.
Return JSON with:
- "target": the answer to the user's question, exactly one of the options it lists.
- "subject": the kind of the main subject.
- "style": the rendering style of the whole image.
- "setting": whether the scene is indoor or outdoor.
- "shot": the camera framing of the main subject.

[User]
Question for "target": {question} Answer with exactly one of: {option_1}; {option_2}.

Also classify:
- subject: one of person, animal, plant, food, object, vehicle,
  building, landscape, abstract
- style: one of photograph, painting, illustration, 3D render, other
- setting: one of indoor, outdoor, unclear
- shot: one of close-up, medium, wide
\end{tcblisting}

\begin{tcblisting}{
  listing only,
  breakable,
  colback=orange!5,
  colframe=orange!40,
  coltitle=orange!60!black,
  title=\textbf{Broken-image judge prompt for component swapping},
  fonttitle=\bfseries,
  boxrule=0.6pt,
  arc=2mm,
  left=1mm,
  right=1mm,
  top=1mm,
  bottom=1mm,
  listing options={
    basicstyle=\ttfamily\scriptsize,
    breaklines=true,
    breakindent=0pt,
    breakautoindent=false,
    columns=fullflexible,
    keepspaces=true,
    showstringspaces=false
  }
}
[System]
You inspect images produced by a text-to-image model. Several images follow the text: all but the last are references, normal unedited outputs of the model for closely related prompts. The LAST image is the output after internal activations were edited. Decide only whether the LAST image still works as a picture.

[User]
The last image may legitimately differ from the references in content, composition, viewpoint, style, color or lighting; such differences are NOT failures. Use the references only to see what an intact output of this model looks like. Judge only the LAST image.
First, in "subject", name the main concrete object or scene you can clearly recognize (for example "a cellist on a stage"), or "none" if you cannot clearly recognize any.
Then classify the image in "verdict":
- "broken": the image does not work as a picture. Use this when no concrete object or scene is clearly recognizable, or when the picture is dominated by noise or static, a mosaic or grid of tiles, smeared texture, a patchwork or collage of unrelated fragments, glitch stripes or blocks, or a nearly flat color, so that only faint remnants of a scene are left. Shapeless smears or textures count as broken even if they look painterly.
- "degraded": a concrete object or scene is still clearly recognizable and forms one coherent picture, but visible artifacts are laid over it, such as a grid or checker pattern, ghosting or a double exposure, or smeared patches.
- "ok": a coherent picture without such artifacts. An unusual composition, viewpoint, style (photograph, any painting style including impressionist or cubist, drawing, pixel art, 3D render, anime), color, black and white, lighting, darkness or crop is NOT a failure, and the content does not have to make sense.
Return JSON with "subject", "verdict" (one of ok, degraded, broken) and "reason" (at most 12 words).
Also answer, for the LAST image, in "target": {choice_question} For reference only: image 1 was generated with a prompt asking for "{v_a}", image 2 was generated with a prompt asking for "{v_b}".
\end{tcblisting}




\section{Additional Results}


\subsection{Other Baselines for Reading Activations}
\label{app:other_baselines_reading}
In Exp.~1 of Sec.~\ref{sec:dao_reading_perf}, we test whether access to prompt information in the DM's activations can explain DAO's advantage over the VLM baseline. We evaluate two additional baselines that receive the generation prompt directly. Llama-3.1-8B-Instruct answers each question using the prompt alone, while Llama-3.2-11B-Vision-Instruct receives both the prompt and the image decoded at the current denoising step.

\begin{figure}[t]
    \centering
    \includegraphics[width=.99\linewidth]{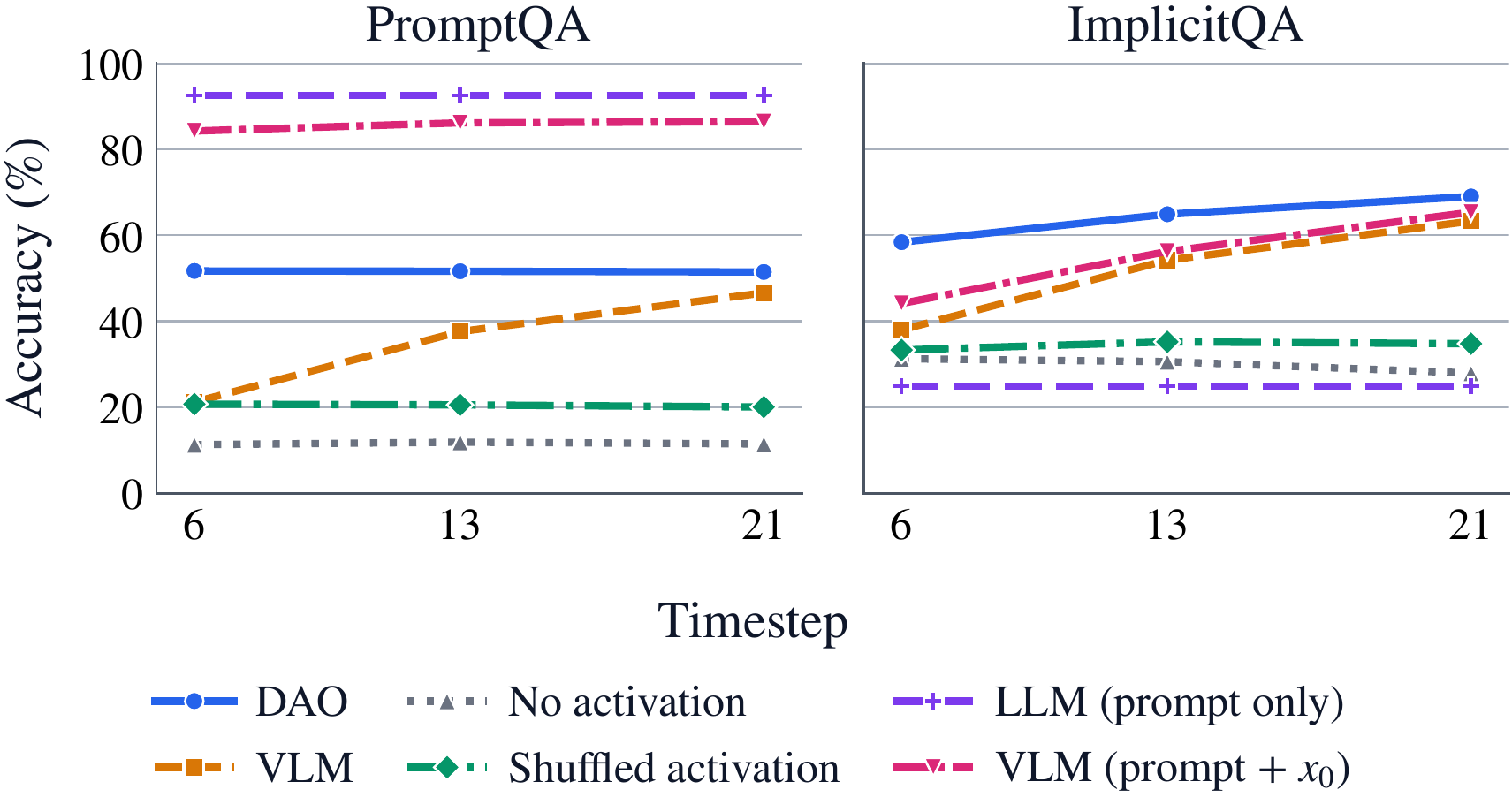}
    \caption{\textbf{Activation reading accuracy across methods with additional baselines.} We compare DAO and the baselines on PromptQA and ImplicitQA across denoising steps. Two additional baselines receive the generation prompt: an LLM given the prompt alone and a VLM given both the prompt and the decoded image.}
    \label{fig:reading_other_baselines}
\end{figure}

As shown in Fig.~\ref{fig:reading_other_baselines}, both baselines achieve approximately 80--90\% accuracy on PromptQA. This is expected because PromptQA asks about details explicitly specified in the prompt, so these scores serve as an oracle reference. On ImplicitQA, DAO outperforms both baselines at every evaluated step, even when the VLM receives both the prompt and the decoded image. This supports the use of ImplicitQA to assess whether DAO can read visual details from the DM's activations beyond those specified in the prompt.

\subsection{Reading Performance on Unseen Timesteps}
\label{app:ood_timestep}

\begin{figure}[t]
    \centering
    \includegraphics[width=.99\linewidth]{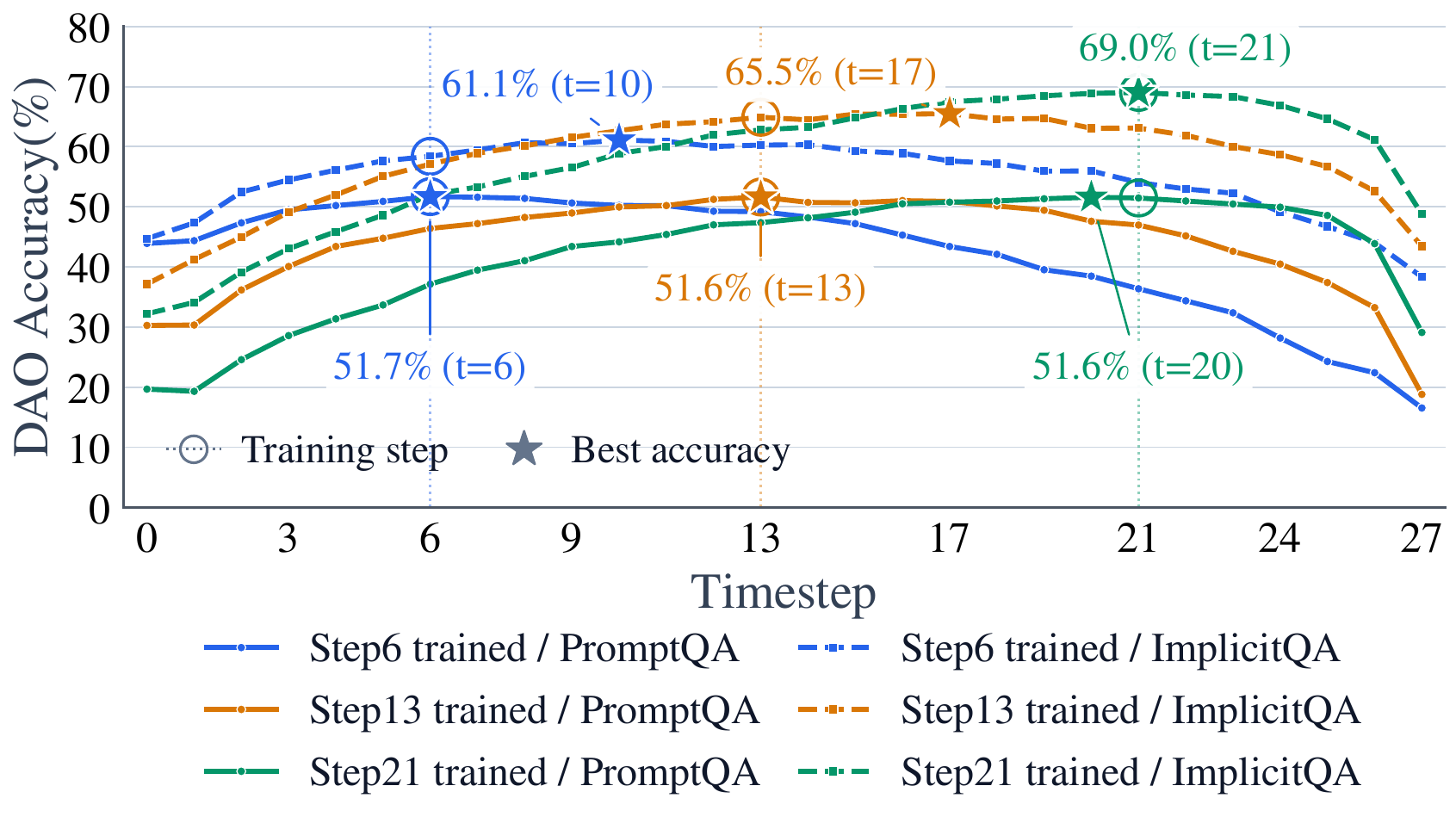}
    \caption{\textbf{DAO generalization to unseen timesteps.}
Blue, orange, and green denote DAOs trained at steps 6, 13, and 21, respectively; solid and dashed lines indicate PromptQA and ImplicitQA.
Circles mark accuracy at the training timestep, and stars mark each curve's peak accuracy.
Accuracy generally peaks near the training timestep and declines farther away, indicating robustness to small timestep shifts but limited generalization across larger ones.}
    \label{fig:reading_ood_timestep}
\end{figure}

We evaluate how the timestep-specific DAOs trained in the main experiments generalize to unseen timesteps.
Following the evaluation in Fig.~\ref{fig:reading_acc}, we measure accuracy on 6,000 PromptQA and ImplicitQA QA pairs for 1,000 unseen activations, using the same LLM-as-judge protocol.

Figure~\ref{fig:reading_ood_timestep} shows the results.
Blue, orange, and green denote DAOs trained at steps 6, 13, and 21, respectively, while solid and dashed lines indicate PromptQA and ImplicitQA.
Across settings, accuracy is generally highest at or near the training timestep, indicating robustness to small timestep shifts.
However, performance declines at more distant timesteps, suggesting limited robustness to the larger shifts in activation distributions associated with greater timestep differences.

\subsection{Reading Accuracy with an Alternative LLM Judge}
\label{app:gemini_judge}

\begin{table}[t]
  \centering
  \caption{\textbf{Accuracy (\%) on PromptQA and ImplicitQA at Step 6, evaluated by Gemini 3.8 Flash.} The last column reports judge agreement (\%) between Gemini and GPT-5.6 Luna on DAO answers.}
  \label{tab:specific_gemini_accuracy}
  \resizebox{\linewidth}{!}{%
    \begin{tabular}{lrrrr|r}
      \toprule
      Task & No act. & Shuffled act. & VLM & DAO & Judge agreement \\
      \midrule
      PromptQA   & 12.20 & 22.20 & 23.97 & \textbf{55.30} & 91.93 \\
      ImplicitQA & 35.87 & 37.53 & 43.50 & \textbf{66.60} & 90.20 \\
      \bottomrule
    \end{tabular}%
  }
\end{table}

In Sec.~\ref{sec:dao_reading_perf}, we used GPT-5.6 Luna both to generate DAO's training data and to judge the answers produced by DAO and the baselines on PromptQA and ImplicitQA.
To control for potential bias from the judge favoring wording similar to that in the training data, we repeated the step 6 evaluation with Gemini 3.8 Flash under the same protocol.
As shown in Table~\ref{tab:specific_gemini_accuracy}, DAO outperforms all baselines on both tasks, consistent with the results in Fig.~\ref{fig:reading_acc}.
The two judges also agreed on whether an answer was correct in over 90\% of cases.
These results suggest that DAO consistently produces accurate descriptions of generated images from intermediate activations and that its performance cannot be attributed solely to the preferences of the model used to generate its training data.

\subsection{Per-attribute Results of Steering}
\label{app:steering_per_attr}
\begin{figure*}[t]
    \centering
    \includegraphics[width=\linewidth]{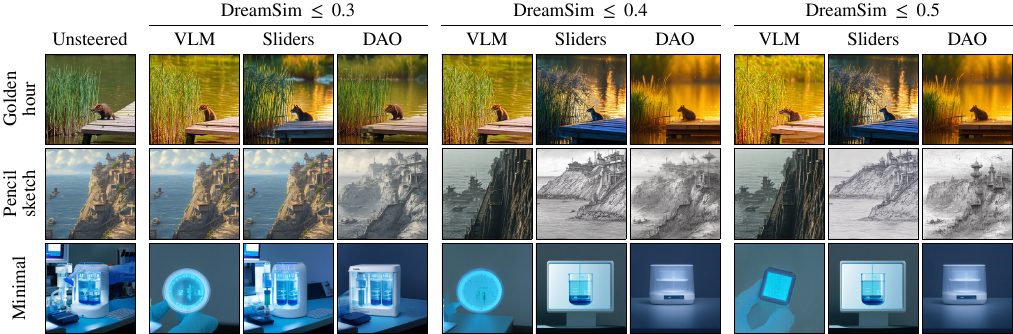}
    \caption{\textbf{Qualitative steering results under all image-change budgets (full version of Fig.~\ref{fig:steering}).}
    Each row shows one held-out prompt and noise, steered toward golden-hour lighting, a pencil-sketch style, and a minimal composition with a plain background, respectively.
    For each budget, the image of each method is generated with the strength selected for that attribute and budget.}
    \label{fig:steering_full}
\end{figure*}

Tab.~\ref{tab:steering_per_attr} lists the per-attribute success rates under matched budgets, and Fig.~\ref{fig:steering_full} shows qualitative results under all three budgets.
Fig.~\ref{fig:steering_full} is the full version of Fig.~\ref{fig:steering}, which shows only the budget of DreamSim $\leq 0.3$.
For each method, attribute, and budget, we use the strength selected as described in Appx.~\ref{app:steering}.
Because the unsteered model is always a candidate, it is selected when no swept strength within the budget increases the success rate; in that case, the image in Fig.~\ref{fig:steering_full} is identical to the unsteered image, as for the VLM and Concept Sliders on pencil sketch under DreamSim $\leq 0.3$.
Note that the strengths are selected and the success rates are reported on the same 32 held-out prompts, for all methods alike.

Under the tightest budget, DAO achieves the highest success rate for seven of the eight attributes; the exception is minimal composition, for which both Concept Sliders and the VLM are higher.
Steering with the VLM succeeds for at most 3\% of the images for ink drawing, anime, and pencil sketch at every budget.
For minimal composition, the VLM-based LoRA replaces the scene with a close-up of a glowing display rather than simplifying the composition (Fig.~\ref{fig:steering_full}).
As the budget is relaxed, Concept Sliders closes the gap for some attributes (e.g., watercolor and rain under DreamSim $\leq 0.4$), but DAO is still the best or tied for the best for at least six of the eight attributes under every budget.

\begin{table}[t]
    \centering
    \caption{\textbf{Per-attribute steering success rate under matched image-change budgets}. Bold indicates the best method for each attribute and budget, including ties.}
    \label{tab:steering_per_attr}
    \resizebox{\linewidth}{!}{%
    \begin{tabular}{lccccccccc}
\toprule
& \multicolumn{3}{c}{DreamSim $\leq$ 0.3} & \multicolumn{3}{c}{DreamSim $\leq$ 0.4} & \multicolumn{3}{c}{DreamSim $\leq$ 0.5} \\
\cmidrule(lr){2-4}\cmidrule(lr){5-7}\cmidrule(lr){8-10}
Attribute & VLM & Sliders & DAO & VLM & Sliders & DAO & VLM & Sliders & DAO \\
\midrule
Watercolor & 0.28 & 0.19 & \textbf{0.84} & 0.41 & \textbf{1.00} & 0.91 & 0.41 & \textbf{1.00} & 0.97 \\
Minimal & 0.41 & \textbf{0.44} & 0.31 & 0.56 & \textbf{0.66} & \textbf{0.66} & \textbf{0.66} & \textbf{0.66} & \textbf{0.66} \\
Sepia & 0.16 & 0.03 & \textbf{0.28} & \textbf{0.69} & 0.66 & \textbf{0.69} & 0.84 & \textbf{0.91} & 0.84 \\
Ink drawing & 0.00 & 0.00 & \textbf{0.34} & 0.00 & 0.00 & \textbf{0.34} & 0.00 & 0.59 & \textbf{0.88} \\
Anime & 0.00 & 0.00 & \textbf{0.22} & 0.00 & 0.38 & \textbf{0.69} & 0.00 & 0.69 & \textbf{0.78} \\
Golden hour & 0.38 & 0.34 & \textbf{0.44} & 0.38 & 0.69 & \textbf{0.75} & 0.44 & 0.69 & \textbf{0.75} \\
Pencil sketch & 0.00 & 0.00 & \textbf{0.22} & 0.03 & 0.50 & \textbf{0.72} & 0.03 & 0.88 & \textbf{0.97} \\
Rain & 0.09 & 0.34 & \textbf{0.59} & 0.12 & \textbf{0.66} & \textbf{0.66} & 0.12 & \textbf{0.66} & \textbf{0.66} \\
\midrule
Mean & 0.164 & 0.168 & \textbf{0.406} & 0.273 & 0.566 & \textbf{0.676} & 0.312 & 0.758 & \textbf{0.812} \\
\#Best & 0 & 1 & \textbf{7} & 1 & 3 & \textbf{7} & 1 & 4 & \textbf{6} \\
\bottomrule
\end{tabular}

    }
\end{table}

\subsection{Per-attribute Results of Component Swapping}
\label{app:swap_per_attr}
\begin{figure*}[t]
    \centering
    \includegraphics[width=\linewidth]{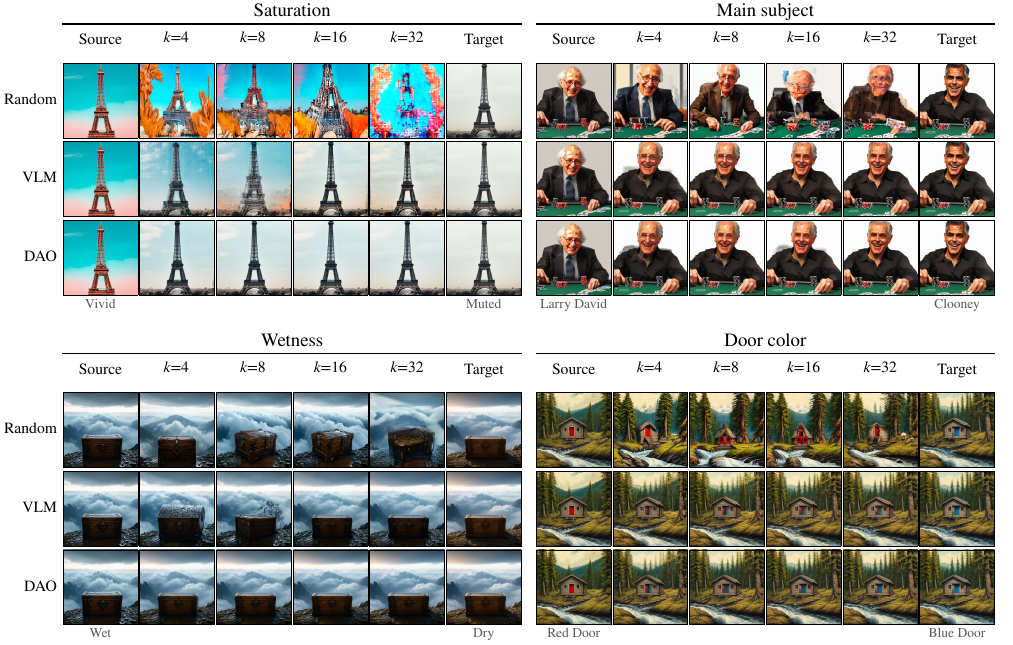}
    \caption{\textbf{Swapping the top-$k$ components identified by each attribution method for saturation, main subject, wetness, and door color.}
    Together with Fig.~\ref{fig:component_swap} (render style), this figure shows all five attributes.
    In each panel, the outputs of the top-$k$ components in the generation of the source image (left) are replaced with those in the generation of the target image (right).
    Rows show components selected at random or by AtP-IG using the VLM (given the decoded $\hat{\vz}_0$) or DAO, and columns show the number of swapped components $k$.}
    \label{fig:component_swap_full}
\end{figure*}

Tab.~\ref{tab:component_swap_per_attr} lists the per-attribute success rates of swapping the top-$k$ components, and Fig.~\ref{fig:component_swap_full} shows qualitative results for the four attributes other than render style.
Together with Fig.~\ref{fig:component_swap}, which shows render style, Fig.~\ref{fig:component_swap_full} forms the full version of the qualitative results for all five attributes.
In each panel, the outputs of the top-$k$ components in the generation of the source image are replaced with those in the generation of the target image; the attribute values of the two images are given below them.

DAO achieves the highest success rate, or ties for it, for every attribute and every $k$.
The VLM approaches DAO for main subject and door color but falls far behind it for render style and wetness; for render style, the VLM succeeds for at most 19\% of the pairs, whereas DAO succeeds for at least 50\%.
Swapping randomly selected components rarely changes the attribute and can break the image at larger $k$, as in the saturation example in Fig.~\ref{fig:component_swap_full}.

\begin{table}[t]
    \centering
    \caption{\textbf{Per-attribute success rate of swapping the top-$k$ components} (16 pairs per attribute). Bold indicates the best method for each attribute and $k$, including ties.}
    \label{tab:component_swap_per_attr}
    \footnotesize
    \resizebox{\linewidth}{!}{\begin{tabular}{llcccc}
\toprule
& & \multicolumn{4}{c}{Number of swapped components} \\
\cmidrule(lr){3-6}
Attribute & Method & $k=4$ & $k=8$ & $k=16$ & $k=32$ \\
\midrule
Render style & Random & 0.188 & 0.188 & 0.312 & 0.188 \\
 & VLM & 0.062 & 0.125 & 0.188 & 0.125 \\
 & DAO (ours) & \textbf{0.500} & \textbf{0.562} & \textbf{0.875} & \textbf{0.875} \\
\midrule
Saturation & Random & 0.000 & 0.062 & 0.000 & 0.000 \\
 & VLM & 0.500 & 0.500 & 0.688 & 0.750 \\
 & DAO (ours) & \textbf{0.688} & \textbf{0.688} & \textbf{0.750} & \textbf{0.938} \\
\midrule
Main subject & Random & 0.062 & 0.125 & 0.062 & 0.250 \\
 & VLM & \textbf{0.500} & \textbf{0.688} & \textbf{0.812} & 0.812 \\
 & DAO (ours) & \textbf{0.500} & \textbf{0.688} & \textbf{0.812} & \textbf{0.938} \\
\midrule
Wetness & Random & 0.125 & 0.062 & 0.062 & 0.188 \\
 & VLM & 0.125 & 0.062 & 0.312 & 0.438 \\
 & DAO (ours) & \textbf{0.625} & \textbf{0.625} & \textbf{0.688} & \textbf{0.688} \\
\midrule
Door color & Random & 0.000 & 0.000 & 0.000 & 0.000 \\
 & VLM & 0.438 & 0.562 & \textbf{0.938} & \textbf{0.875} \\
 & DAO (ours) & \textbf{0.500} & \textbf{0.688} & \textbf{0.938} & \textbf{0.875} \\
\bottomrule
\end{tabular}
}
\end{table}

\subsection{Steering the Identified Components}
\label{app:component_steer}
To establish that the identified components are specifically important for the target attribute rather than merely for image generation in general, we steer only the top-8 components and compare the results with steering the top-8 components from the analysis of another attribute and eight randomly selected components.
If steering the top components for the target attribute succeeds more often than the other settings, these components are particularly important for the attribute, and AtP-IG with DAO is useful for analyzing the roles of components.

\paragraph{Components.}
The target components are the top-8 components of the attribute at step 6.
The components for another attribute (Other) are the top-8 components of the attribute whose top-8 components overlap the least with the target components; this was main subject for all four target attributes.
The random components (Random) are drawn so that each matches the kind of the corresponding target component.

\paragraph{Training.}
We use the DAO-based steering method described in Sec.~\ref{sec:steering}, attaching the LoRA only to the selected components.
Since the steering method is shared across the three settings and this experiment compares how strongly LoRA training on each set of components can express the attribute change, rather than comparing steering methods, we fix the strength to $\gamma=1$, the strength used during training.

\paragraph{Evaluation.}
For each attribute, we evaluate one direction for which the success rate without steering is below 0.2: door red $\to$ blue, dry $\to$ wet, vivid $\to$ grayscale, and photograph $\to$ oil painting.
In the opposite directions of wetness, saturation, and render style, the target value already appears in most unsteered images.
Main subject is not steered because its values differ from pair to pair (Tab.~\ref{tab:atp_attrs}).
An LLM generates 100 prompts per attribute that contain the target object but do not specify the attribute, and we generate two images per prompt with different noise.
The success rate is computed as in Sec.~\ref{sec:steering}, using the judge prompt in Appx.~\ref{app:steering} with \texttt{\{question\}} replaced by the judge question in Tab.~\ref{tab:component_steer_q}.

\begin{table}[t]
    \centering
    \caption{\textbf{Judge questions for component steering.}}
    \label{tab:component_steer_q}
    \footnotesize
    \begin{tabular}{l p{0.6\linewidth}}
    \toprule
    Direction & Judge question \\
    \midrule
    Door: red $\to$ blue & \textit{Is the door of the building in this image blue?} \\
    Dry $\to$ wet & \textit{Is the \{subject\} in this image wet?} \\
    Vivid $\to$ grayscale & \textit{Is this image in grayscale (black and white)?} \\
    Photo $\to$ oil painting & \textit{Does this image look like an oil painting?} \\
    \bottomrule
    \end{tabular}
\end{table}

\begin{table}[t]
    \centering
    \caption{\textbf{Steering only the selected components at $\gamma=1$.}
    Success is the fraction of images for which the judge answers \textit{yes} to whether the image has the target value (200 images per direction).}
    \label{tab:component_steer}
    \footnotesize
    \setlength{\tabcolsep}{4pt}
\begin{tabular}{lccc}
\toprule
& \multicolumn{3}{c}{Success $\uparrow$} \\
\cmidrule(lr){2-4}
Direction & Random & Other & Target \\
\midrule
Door: red $\to$ blue & 0.095 & 0.245 & \textbf{0.330} \\
Dry $\to$ wet & 0.040 & 0.065 & \textbf{0.120} \\
Vivid $\to$ grayscale & 0.000 & 0.090 & \textbf{0.330} \\
Photo $\to$ oil painting & 0.190 & 0.275 & \textbf{0.595} \\
\midrule
Average & 0.081 & 0.169 & \textbf{0.344} \\
\bottomrule
\end{tabular}

\end{table}

\begin{figure*}[t]
    \centering
    \includegraphics[width=\linewidth]{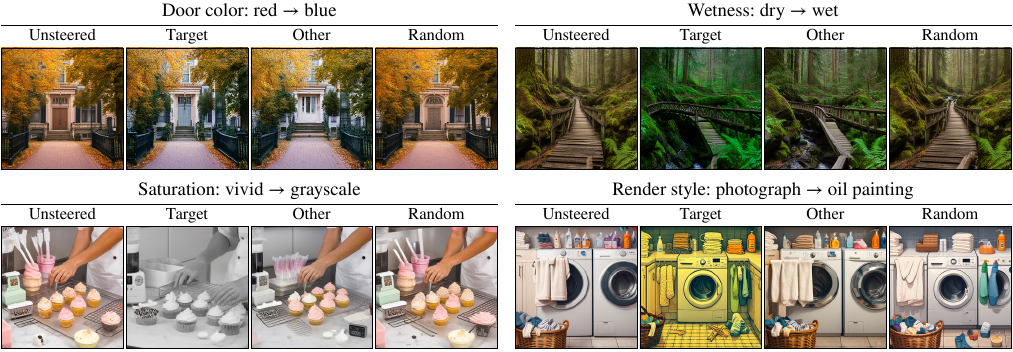}
    \caption{\textbf{Qualitative results of steering only the selected components at $\gamma=1$.}
    Each panel shows one steering direction for one prompt and noise.
    Unsteered is the image without steering, and Target, Other, and Random denote steering the top-8 components for the target attribute, the top-8 components for another attribute, and eight randomly selected components, respectively.}
    \label{fig:component_steer}
\end{figure*}

\paragraph{Results.}
Tab.~\ref{tab:component_steer} shows the success rates, and Fig.~\ref{fig:component_steer} shows qualitative results for one prompt and noise per direction.
Steering the target components succeeds most often for all four directions.
The success rate for dry $\to$ wet is low in all three settings, but the target components still give the highest rate.
In Fig.~\ref{fig:component_steer}, steering the target components turns the door blue, the image grayscale, and the photograph into an oil painting, whereas steering the random components leaves the images almost unchanged, and steering the components for the other attribute changes the images in ways unrelated to the target value (e.g., the door becomes white rather than blue).
This indicates that AtP-IG extracts components that are sensitive to manipulations of the attribute.
In other words, AtP-IG with DAO can identify not only components that are important for image generation in general but also the components of the DM that are important for the attribute specified in natural language through DAO.

\subsection{Block-Level Analysis of Component Importance by Attribute}
\label{app:atp_blockwise}
\begin{figure*}[t]
    \centering
    \includegraphics[width=\linewidth]{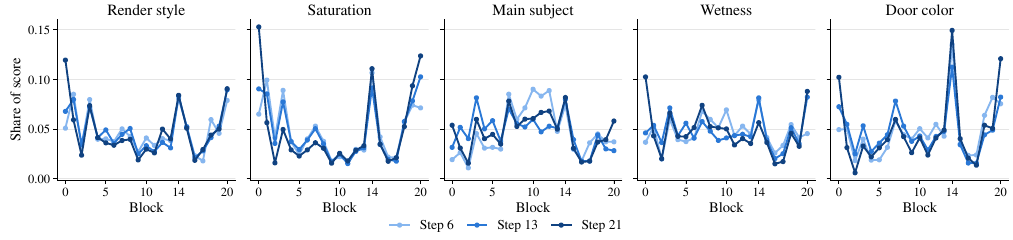}
    \caption{\textbf{Block importance estimated by AtP-IG with DAO.} For render style, saturation, main subject, wetness, and door color, we plot each block's share of the total positive attribution score at denoising steps 6, 13, and 21.}
    \label{fig:component_blocks}
\end{figure*}

In Sec.~\ref{sec:attribution}, we use AtP-IG with DAO to score individual components for each target attribute. Here, we aggregate these scores by block to examine how block importance varies across attributes and denoising steps. For each attribute at steps 6, 13, and 21, Fig.~\ref{fig:component_blocks} plots the block index on the horizontal axis and the fraction of the total positive attribution score accounted for by that block on the vertical axis.

For door color, a localized visual attribute, Block~14 accounts for the largest share at all three steps, suggesting that it may help determine fine-grained details such as the color of an individual object. For image-wide attributes such as render style and saturation, Block~14 also contributes at all three steps, but Block~0's share increases over time and exceeds that of Block~14 by step~21. For main subject and wetness, Block~0 and~14 together account for a smaller share than for the other three attributes at every evaluated step. Both attributes involve cues at multiple scales, from object placement to surface texture for main subject, and from overall color tone to individual raindrops for wetness. These patterns suggest that the relative importance of each block depends on the target attribute.

\section{Relation to a Contemporaneous Work}



At a late stage of our research, we became aware of \citet{dahary2026learning}, who also use LLMs to verbalize the internal activations of DMs.
Their paper was released on arXiv on October 5, 2026, and we regard it as contemporaneous work.
Their work investigates the role of text-side tokens in MM-DiT through multimodal attention, using LLM-based readout to understand their representations.
They find that these tokens contain substantial information about the final generated image even at early denoising steps, suggesting that they acquire visual semantic information through multimodal attention.
They also observe a positive association between the readability of these representations, measured by LLM-based QA performance, and the quality of the final generated images.
Motivated by this finding, they demonstrate that encouraging contextual representations to encode richer visual semantic information during DM training can improve generation performance.

The two studies differ in the following respects.

\paragraph{Conceptual Framework.}
Their work does not explicitly connect its approach to LatentQA or Activation Oracles.
Our work explicitly extends this framework to DMs.

\paragraph{Applications.}
Based on their readability analysis, they propose Contextual Alignment (CoAl), a training method for improving DM generation quality.
The reader serves as an analysis tool that motivates CoAl's design, while CoAl itself operates independently of the reader.
Our work directly employs DAO for reading internal representations, steering generation, and identifying important model components, enabling model control and interpretation through a natural-language interface.

\paragraph{Task Design.}
Their reader is trained and evaluated on questions about images featuring a single person.
Our work considers a broader range of visual content, including people, animals, objects, landscapes, and artistic styles.
We also distinguish between PromptQA and ImplicitQA based on whether the queried information is explicitly specified in the generation prompt, allowing us to evaluate the two types of information separately.

\paragraph{Architecture.}
Their reader aggregates text-side contextual tokens from all MM-DiT layers, specifically the outputs at text-token positions immediately after attention, and projects them through a Q-Former before injecting them into an LLM.
Their training focuses on the projection module, whereas our approach jointly trains $G_\phi$ and LoRA adapters in the decoder LLM.

\paragraph{Training Scale.}
They use 150K QA pairs derived from 2K generation prompts for training, whereas our approach uses approximately 2.5M QA pairs derived from 500K prompts.

We emphasize that these distinctions are intended to clarify the differences between the two studies rather than to claim the superiority of either approach. We find this contemporaneous work highly interesting and novel, and we hope that both studies will encourage further exploration of natural-language interpretation of internal activations in DMs.


\end{document}